\documentclass[10pt,twocolumn,letterpaper]{article}

\usepackage[pagenumbers]{iccv}      % To produce the REVIEW version
\usepackage{amsmath,amsfonts,bm}

\def\eqref#1{Eq.~(\ref{#1})}
\def\1{\bm{1}}

\DeclareMathAlphabet{\mathsfit}{\encodingdefault}{\sfdefault}{m}{sl}
\SetMathAlphabet{\mathsfit}{bold}{\encodingdefault}{\sfdefault}{bx}{n}

\usepackage{graphicx}
\usepackage{amsmath}
\usepackage{amssymb}
 \usepackage{multirow}
\usepackage{booktabs}
\usepackage{algorithm}
\usepackage{algpseudocode}
\usepackage{amsthm} % for theorem environment
\usepackage{thmtools}
\usepackage{graphicx} % for images, if needed
\usepackage{booktabs}
\usepackage{pdflscape}
\usepackage{adjustbox}
\usepackage{multibib}
\usepackage{longtable}
\newcites{app}{Appendix References}
\def\paperID{6701}

\newtheorem{Definition}{Definition}

\newtheorem{theorem}{Theorem}

\usepackage{dsfont}
\usepackage[pagebackref,breaklinks,colorlinks]{hyperref}
\usepackage{array}

\usepackage[capitalize]{cleveref}
\Crefname{section}{Section}{Sections}
\crefname{figure}{Figure}{Figures}
\crefname{table}{Table}{Tables}
\crefname{tabular}{Table}{Tables}

\def\confName{ICCV}
\def\confYear{2025}

\begin{document}

% \title{Extracting Training Data from Unconditional Diffusion Models}
\title{SIDE: Surrogate Conditional Data Extraction from Diffusion Models}

\author{
 Yunhao Chen \\
 Fudan University \\
  {\texttt{24110240013@m.fudan.edu.cn}} \\
  \and
  Shujie Wang \\
 Fudan University \\
  {\texttt{24110240084@m.fudan.edu.cn}} \\
  \and
  Difan Zou \\
  University of Hong Kong \\
  {\texttt{dzou@cs.hku.hk}} \\
  \and
    Xingjun Ma \\
  Fudan University \\
  {\texttt{xingjunma@fudan.edu.cn}} \\
}
\maketitle

%%%%%%%%% ABSTRACT
\begin{abstract}
% As diffusion probabilistic models (DPMs) become the mainstream framework for Generative AI (GenAI), understanding their memorization behavior is crucial for assessing risks related to data leakage, copyright infringement, and trustworthy deployment. Previous studies suggest that conditional DPMs are more susceptible to memorization than their unconditional counterparts, and most existing extraction techniques are designed for conditional models. 
% However, studying unconditional DPMs remains essential, as they underpin advanced systems like Stable Diffusion. Gaining insights into their memorization patterns can illuminate broader characteristics of DPMs as a whole.
% In this work, we introduce \textbf{Surrogate condItional Data Extraction (SIDE)}, a novel method for extracting training data from unconditional DPMs. SIDE leverages a time-dependent classifier, trained on generated samples, to serve as surrogate conditions for data extraction. Empirical results demonstrate that SIDE can effectively extract training data in challenging scenarios where existing methods fail, achieving an average performance gain of over 50\% on the CelebA dataset across various scales.
% Furthermore, we present a unified theoretical framework that explains both why SIDE is effective and how memorization manifests in conditional and unconditional DPMs. Our findings advance the understanding of memorization in diffusion models and offer practical guidance for mitigating risks.

As diffusion probabilistic models (DPMs) become central to Generative AI (GenAI), understanding their memorization behavior is essential for evaluating risks such as data leakage, copyright infringement, and trustworthiness. While prior research finds conditional DPMs highly susceptible to data extraction attacks using explicit prompts, unconditional models are often assumed to be safe. We challenge this view by introducing \textbf{Surrogate condItional Data Extraction (SIDE)}, a general framework that constructs data-driven surrogate conditions to enable targeted extraction from any DPM. Through extensive experiments on CIFAR-10, CelebA, ImageNet, and LAION-5B, we show that SIDE can successfully extract training data from so-called safe unconditional models, outperforming baseline attacks even on conditional models. Complementing these findings, we present a unified theoretical framework based on informative labels, demonstrating that all forms of conditioning, explicit or surrogate, amplify memorization. Our work redefines the threat landscape for DPMs, establishing precise conditioning as a fundamental vulnerability and setting a new, stronger benchmark for model privacy evaluation.
\end{abstract}

%%%%%%%%% BODY TEXT
% For example, \cite{Gu2023OnMI} found that conditioning on random-labeled data significantly increases memorization, while unconditional models exhibit much lower memorization rates. 

% Despite these findings, investigating data extraction attacks on unconditional DPMs remains crucial for two reasons: (1) unconditional DPMs form the foundation of conditional models like Stable Diffusion, which adapt pre-trained unconditional DPMs for conditional generation \cite{rombach2022high,zhang2023adding}; and (2) understanding memorization in unconditional DPMs provides insights into the broader memorization mechanisms of DPMs.

% In this work, we propose Surrogate condItional Data Extraction (SIDE), a novel method for extracting training data from unconditional DPMs. SIDE employs a time-dependent classifier trained on generated data as surrogate conditions to effectively extract training samples. We empirically validate SIDE on CIFAR-10, multiple scales of the CelebA and ImageNet, demonstrating its superiority over baseline methods by achieving over 50\% higher effectiveness on average. Examples of extracted images are shown in \cref{fig:results_show}.
\section{Introduction}

Diffusion probabilistic models (DPMs) \cite{ho2020denoising,sohl2015deep,song2019generative} are a powerful class of generative models that learn data distributions by progressively corrupting data through a forward diffusion process and then reconstructing it via a reverse process. Owing to their remarkable ability to model complex data distributions, DPMs have become the foundation for many leading Generative Artificial Intelligence (GenAI) systems, including Stable Diffusion \cite{rombach2022high}, DALL-E 3 \cite{BetkerImprovingIG}, and Sora \cite{videoworldsimulators2024}.

However, the widespread adoption of DPMs has raised concerns about \emph{data memorization}, which is the tendency of models to memorize raw training samples. This can lead to the generation of duplicated rather than novel content, increasing the risks of data leakage, privacy breaches, and copyright infringement~\cite{Somepalli2022DiffusionAO, somepalli2023understanding,asay2020independent,cooper2024files}. For example, Stable Diffusion has been criticized as a ``21st-century collage tool” for remixing copyrighted works of artists whose data was used during training~\cite{butterick2023stable}. Furthermore, memorization can facilitate data extraction attacks, enabling adversaries to recover training data from deployed models. Recent work by~\cite{carlini2023extracting,webster2023reproducible} demonstrated the feasibility of extracting training data from DPMs such as Stable Diffusion~\cite{rombach2022high}, highlighting substantial privacy and copyright risks.

Existing studies show that conditional DPMs are far more prone to memorizing training data than unconditional ones, making extraction from unconditional models extremely challenging~\cite{somepalli2023understanding,Gu2023OnMI}. While conditional models can be compromised via prompts, unconditional models are generally seen as much safer, and current extraction methods struggle without detailed prompts.

\begin{figure}
    \centering
    \includegraphics[width=0.45\textwidth]{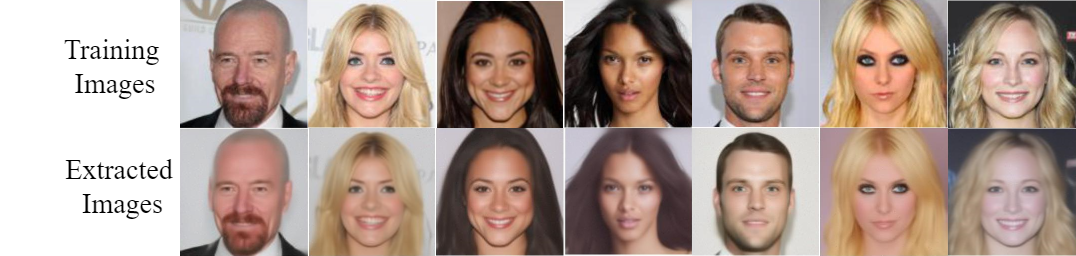}
    \caption{Examples of training images (top) and corresponding extracted images by our SIDE method (bottom) from a DDPM trained on a subset of CelebA. }
    \label{fig:results_show}
   
\end{figure}

To bridge this gap, we propose \textbf{Surrogate condItional Data Extraction (SIDE)}, a general and effective approach for extracting training data from both conditional and unconditional DPMs. SIDE uses cluster information on generated images as a surrogate condition, providing precise guidance toward target samples. This approach outperforms conventional text prompts/class index for conditional models and enables robust extraction attacks on unconditional models. Examples of extracted images are shown in \cref{fig:results_show}.
Additionally, we introduce a divergence measure to quantify memorization in DPMs and provide a theoretical analysis that explains: (1) why conditional DPMs are more susceptible to memorization, even with random labels, and (2) why SIDE is effective for data extraction.

In summary, our main contributions are as follows:
\begin{itemize}
    \item We propose \textbf{SIDE}, a novel data extraction method that leverages a surrogate condition to extract training data from DPMs.
    
    \item We introduce a divergence-based memorization measure and provide a theoretical analysis of the impact of conditioning in DPMs and the effectiveness of SIDE.
    
    \item Experiments on CIFAR-10, CelebA, ImageNet, and LAION-5B show that SIDE can extract training data from unconditional DPMs, often with even greater efficacy than attacks on conditional counterparts, offering new perspectives on the privacy risks of DPMs.
\end{itemize}

\begin{figure*}
    \centering
    \includegraphics[width=1.00\linewidth]{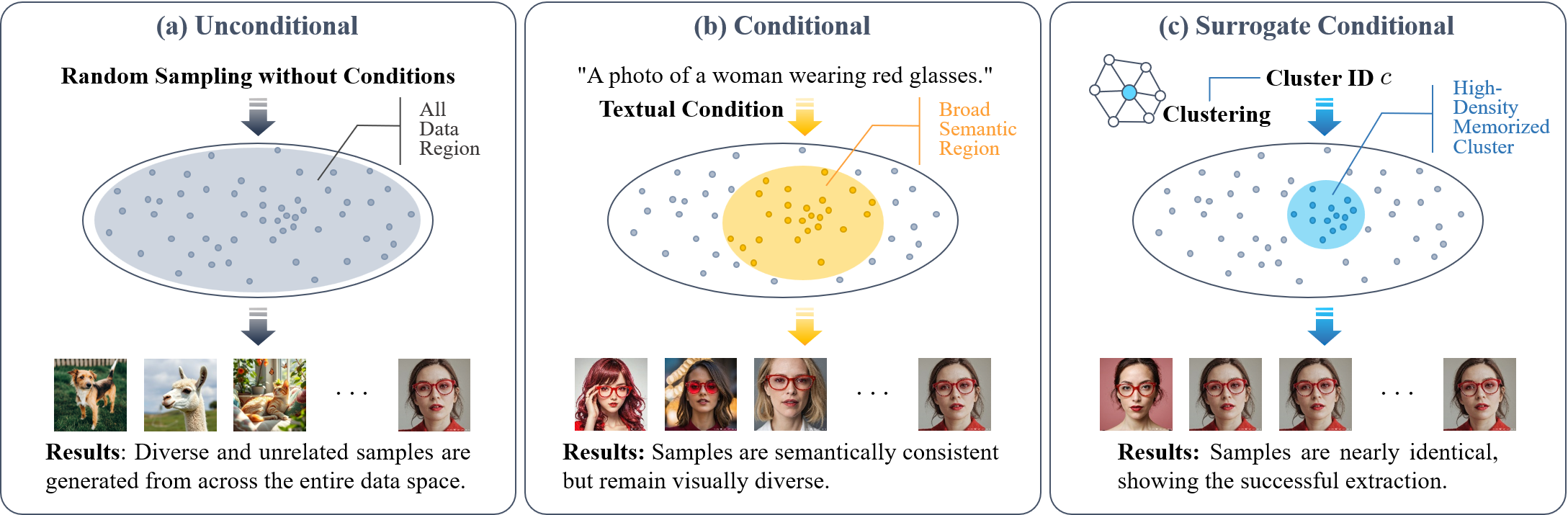}
    \caption{Rationale behind SIDE’s effectiveness. Compared to unconditional models (a), conditional models (b) tend to memorize more due to prompt-based semantic guidance, but this guidance remains too broad for reliable extraction. Our SIDE (c) overcomes this by identifying high-density memorized clusters and creating precise surrogate conditions, enabling more accurate and direct extraction from unconditional models than is possible with conventional conditional approaches. }%\vspace{-1em}}
    
    \label{fig:intuition}
\end{figure*}

% In this work, we first introduce a memorization metric to quantify the degree of memorization in DPMs by measuring the overlap between the generated and training data in a point-wise manner. Based on this metric, we present a theoretical framework that explains why conditional generative models memorize more data and why random labeling can lead to increased memorization. Our theoretical analysis indicates that a classifier trained on the same or similar training data can serve as a surrogate condition for unconditional DPMs. By further making the classifier time-dependent (to the diffusion sampling process), we propose a novel data extraction method named \textbf{Surrogate condItional Data Extraction (SIDE)} to extract training data from unconditional DPMs. We empirically verify the effectiveness of SIDE on CIFAR-10 and different scales of the CelebA dataset (attack results in Figure \ref{fig:results_show}), confirming the accuracy of the theoretical framework.
\section{Related Work}
\paragraph{Diffusion Probabilistic Models.}
DPMs~\cite{sohl2015deep} have achieved state-of-the-art performance in image and video generation, as exemplified by models such as Stable Diffusion~\cite{rombach2022high}, DALL-E 3~\cite{BetkerImprovingIG}, Sora~\cite{videoworldsimulators2024}, Runway~\cite{rombach2022high} and Imagen~\cite{saharia2022photorealistic}. These models excel on various benchmarks~\cite{dhariwal2021diffusion}. DPMs can be interpreted from two perspectives: 1) \emph{score matching}~\cite{song2019generative}, where model learns the gradient of data distribution~\cite{song2020score}, and (2) \emph{denoising diffusion}~\cite{ho2020denoising}, where Gaussian noise is added to clean images over multiple time steps, and the model is trained to reverse this process. For conditional sampling, \cite{dhariwal2021diffusion} introduced classifier guidance to steer the denoising process, while \cite{ho2022classifier} proposed classifier-free guidance, enabling conditional generation without explicit classifiers.

\paragraph{Memorization in Diffusion Models.}
Early research on memorization primarily focused on language models~\cite{carlini2022quantifying,jagielski2022measuring}, which later inspired subsequent studies on DPMs~\cite{somepalli2023understanding,dar2024unconditional, Gu2023OnMI,rahman2024frame,shah2025does,dutt2025devil,achilli2025memorization,dutt2025devil,liu2025copyjudge,wu2025taking,baptista2025memorization,fang2024understanding,kowalczuk2025finding,dhanuka2025magic,zheng2025rg3,garnier2025early,hintersdorfunderstanding,zeno2025diffusion,jeonunderstanding,fang2025closer,brokman2025identifying,lyu2025resolving,favero2025bigger,chen2025enhancing,li2024loyaldiffusion,halder2024memorization,jiang2025image}, from quantifying direct data duplication~\cite{Somepalli2022DiffusionAO, carlini2023extracting} to inferring the presence of an entire identity within the training data~\cite{vora2025identity}.  Notably, \cite{Somepalli2022DiffusionAO} found that 0.5-2\% of generated images duplicate training samples, a result corroborated by \cite{carlini2023extracting} through more extensive experiments on both conditional and unconditional DPMs. Further studies~\cite{somepalli2023understanding, Gu2023OnMI} linked memorization to model conditioning, showing that conditional DPMs are more prone to memorization.
To address memorization, several methods have been proposed for detection and mitigation. For example, \citet{wen2024detecting} introduced a method to detect memorization-triggering prompts by analyzing the magnitude of text-conditional predictions, achieving high accuracy with minimal computational overhead. \citet{ren2024unveiling} proposed metrics based on cross-attention patterns in DPMs to identify memorization. On the mitigation side, \citet{chen2024towards} developed anti-memorization guidance to reduce memorization during sampling, while \citet{ren2024unveiling} modified attention scores or masked summary tokens in the cross-attention layer. \citet{wen2024detecting} minimized memorization by controlling prediction magnitudes during inference.

Despite recent advances, the effectiveness and focus of current research on data extraction have been uneven. Most successful attacks target conditional DPMs, leveraging explicit conditions (e.g., prompts) to guide the generation process toward memorized samples~\cite{carlini2023extracting,wu2024leveraging}. In contrast, extracting data from unconditional DPMs has proven to be significantly more challenging due to the absence of such guidance mechanisms~\cite{Gu2023OnMI}. 
To gain deeper insight into memorization in both conditional and unconditional DPMs, we introduce a novel and general data extraction method that enables effective extraction across both model types.
% By creating a powerful surrogate condition, SIDE provides the targeted guidance necessary for effective extraction from unconditional models, offering a more precise alternative to text prompts for conditional models. We complement our method with a theoretical framework explaining memorization across both conditional and unconditional DPMs.
% Despite these advances, the current research mostly focuses on conditional DPMs~\cite{carlini2023extracting,webster2023reproducible}, as extracting data from unconditional DPMs has been found to be extremely more challenging~\cite{Gu2023OnMI}. However, investigating unconditional DPMs remains critical because 1) they form the foundation of conditional models like Stable Diffusion, which adapt them for conditional generation, and 2) understanding memorization in unconditional DPMs provides insights into the broader behavior of DPMs.
% In response to these challenges, we propose \textbf{SIDE}, a novel data extraction method for unconditional DPMs, and provide a unified theoretical framework to explain memorization in both conditional and unconditional DPMs.

\section{Surrogate Conditional Data Extraction}
% This section introduces our proposed Surrogate condItional Data Extraction (\textbf{SIDE}) method for extracting training data from unconditional DPMs.

\paragraph{Threat Model.}
We adopt a white-box threat model in which the attacker has full access to the model parameters. The attacker’s goal is to extract original training samples from the target DPM, whether it is conditional or unconditional. \textbf{In the Appendix, we further extend our SIDE method to black-box and backdoor scenarios.}

% Notably, as we empirically demonstrate in \cref{para:no-leakage}, the classifier itself does not leak training data; its sole purpose is to provide implicit labels, which could alternatively be obtained through clustering techniques.

\subsection{Intuition of SIDE}
Conditional DPMs are known to be more prone to memorization because they rely on explicit labels, such as class tags or prompts, that help steer the model toward specific samples~\cite{Gu2023OnMI, somepalli2023understanding}. 
% Unconditional DPMs, on the other hand, lack these explicit labels but still learn to organize their training data into clusters, even if these groupings are never formally defined~\cite{chen2024deconstructing}.
Unconditional DPMs, by contrast, are trained without explicit labels, yet they implicitly partition the training data into latent clusters, even though these groupings are never explicitly specified~\cite{chen2024deconstructing}.
We refer to these as \textbf{implicit labels}.

The key intuition behind SIDE is that if we can uncover and formalize these clustering patterns within the training data, we can effectively “create” implicit labels to enable conditional control over the model’s outputs.. This approach is powerful because it harnesses the model’s own internal structure for guidance, providing a more direct and targeted way to reach memorized samples than traditional extraction techniques (see Figure~\ref{fig:intuition}).
Below, we outline how to construct implicit labels for unconditional DPMs.

\subsection{Constructing Implicit Labels}
\begin{figure*}[htbp]
    \centering
    \includegraphics[width=0.9\textwidth]{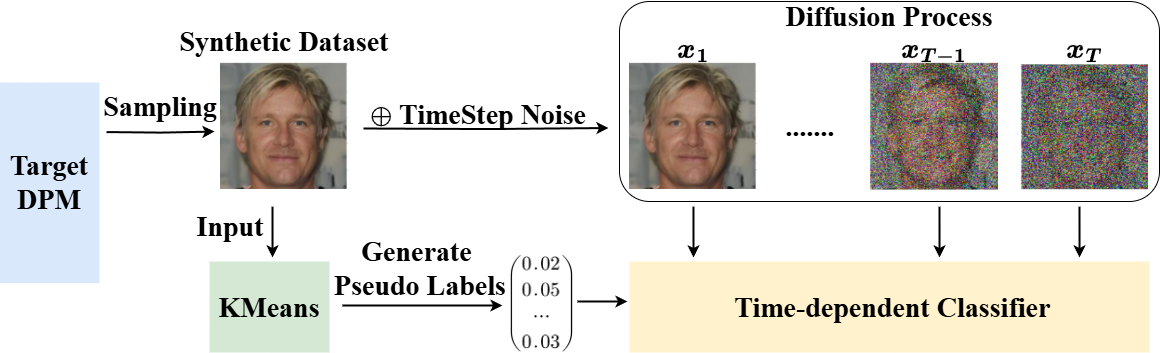} % Replace with the correct file path
    \caption{An illustration of time-dependent classifier training on a pseudo-labeled synthetic dataset.}
    \label{data_distil}
    
\end{figure*}

To generate implicit labels without access to the original training data, we cluster a set of generated images using a pre-trained feature extractor. Clusters with low cohesion (measured by cosine similarity) are removed, and the centroids of the remaining high-quality clusters serve as our surrogate conditions, $y_I$. These conditions guide the DPM's reverse sampling process toward specific, high-density regions where memorized data is likely to reside. Although this guidance can be implemented via a gradient term $\nabla_x \log p_{\theta}^{t}(y_I|x)$, neural classifiers are often miscalibrated. To address this, we introduce a hyperparameter $\lambda$ to adjust the guidance strength, resulting in our final SDE:
\begin{align}
\label{eq:sampling}
\mathrm{d}x= &\Big[ f(x,t)-g(t)^2\Big( \nabla_ x\log p_{\theta}^{t}(x)+ \notag \\
&\nabla_x\log p_{\theta}^{t}(y_I |x) \Big) \Big] \mathrm{d}t+g(t)\mathrm{d}w.
\end{align}
Our formulation, grounded in a power prior, offers a more principled justification for classifier guidance with $\lambda \neq 1$ than previous work~\cite{dhariwal2021diffusion}. The process for training the time-dependent classifier $p_{\theta}^{t}(y_I|x)$ on a pseudo-labeled synthetic dataset is illustrated in Figure~\ref{data_distil}.

\subsection{Training with Surrogate Conditions}
\label{sec:surrogate}
To guide the diffusion model toward class-specific data, we first establish a conditional generation mechanism using pseudo-labels. We explore two distinct approaches for creating these surrogate conditions, selecting the method based on the architecture and scale of the target DPM. For large-scale models like Stable Diffusion, we use parameter-efficient LoRA fine-tuning. For smaller diffusion models, we adopt the traditional approach of training an external, time-dependent classifier for guidance. Both methods begin by generating a synthetic dataset with the target DPM and assigning pseudo-labels via feature clustering with a pre-trained extractor, following established techniques~\cite{chen2023data,chen2024comprehensive}.

\paragraph{Method 1: Training a Time-Dependent Classifier for Small-scale DPMs.}
For small-scale diffusion models, we train an external, time-dependent classifier. Given each synthetic image $x$ and its pseudo-label $y$, we simulate the forward diffusion process by adding Gaussian noise at various timesteps $t$, producing a set of noisy samples $(x_t, t, y)$. The classifier architecture is adapted to accept the timestep $t$ as input (see Figure~\ref{resnet_time} in the Appendix), and is trained on this noisy dataset. The goal is to predict the original label $y$ from the noisy image $x_t$ by minimizing:
\begin{align}
\label{loss_classifier_train}
\mathcal{L}_{\text{cls}} = \mathbb{E}_{t, (x_t, y) \sim \mathcal{D}_{\text{noisy}}}[-\log p_{\theta}^t(y |x_t)]
\end{align}
This training process is illustrated in Figure~\ref{data_distil}.
The resulting classifier $p_{\theta}^t(y|x_t)$ provides an external guidance signal during the reverse diffusion process.

\paragraph{Method 2: LoRA Fine-tuning for Large-scale DPMs.}
For large-scale models such as Stable Diffusion, training a separate classifier is computationally intensive. Instead, we leverage LoRA~\cite{hu2021lora} to directly fine-tune the DPM. Specifically, we freeze the original DPM parameters and insert trainable, low-rank matrices into the U-Net architecture. These lightweight adapters are then fine-tuned on our synthetic dataset, conditioning the DPM on the pseudo-labels $y$. The training objective is to minimize the standard diffusion loss with conditioning:
\begin{align}
\label{loss_lora_train}
\mathcal{L}_{\text{LoRA}} = \mathbb{E}_{t, x_0, \epsilon, y}[|\epsilon - \epsilon_{\theta + \Delta\theta}(x_t, t, y)|^2]
\end{align}
where $\theta$ denotes the frozen DPM weights and $\Delta\theta$ are the trainable LoRA parameters.

\subsection{Overall Procedure of SIDE}
Our \textbf{SIDE} method comprises two main phases. First, it generates a synthetic dataset and assigns pseudo-labels to establish a surrogate guidance mechanism, training the conditional model with the appropriate method described above. During extraction, SIDE applies guidance at each denoising step to steer $x_t$ toward a randomly selected target cluster—using a classifier gradient for small-scale DPMs or conditioning via the LoRA-adapted model for large-scale DPMs.  We then evaluate SIDE using similarity scores on the extracted images and introduce comprehensive metrics for robust assessment in our experiments.

% distribution-based performance metric and provide a theoretical analysis of memorization in DPMs.
\begin{algorithm}
\caption{Surrogate Conditional Data Extraction}
\label{alg:side}
\small
\begin{algorithmic}[1]
  \Require DPM $s_{\theta}(\boldsymbol{x}_t,t)$; feature extractor $F(\cdot)$; clusters $K$; guidance scale $\lambda$; LoRA rank $r$; generations $N_G$; synthetic samples $N_{\text{syn}}$; timesteps $T$; denoiser $DS(\cdot)$; cohesion threshold $\tau$
  \Ensure Extracted data $\mathcal{D}_{\text{ext}}$

  \Statex
  \State \textbf{Part 1: Train Surrogate Conditional Model}
  \State \textit{// Step 1: Generate labeled synthetic dataset}
  \State Generate synthetic data $\mathcal{D}_{\text{img}} = \{\boldsymbol{x}_0^{(i)}\}_{i=1}^{N_{\text{syn}}}$ where $\boldsymbol{x}_0^{(i)} \sim s_\theta$.
  \State Extract features $\mathcal{Z} = \{F(\boldsymbol{x}_0^{(i)}) \mid \boldsymbol{x}_0^{(i)} \in \mathcal{D}_{\text{img}}\}$.
\State $\{\mathcal{C}_k, \mu_k\}_{k=1}^K \gets \text{KMeans}(\mathcal{Z}, K)$ \Comment{Get clusters and centroids}
    \State $\{\mu_k\}_{k=1}^{K'} \gets \left\{\mu_k \mid \frac{1}{|\mathcal{C}_k|} \sum_{\boldsymbol{z} \in \mathcal{C}_k} \frac{\boldsymbol{z} \cdot \mu_k}{\|\boldsymbol{z}\| \|\mu_k\|} \ge \tau \right\}$
  \State Assign labels $y^{(i)} = \operatorname{argmin}_{k \in \{1,\dots,K'\}} \text{dist}(F(\boldsymbol{x}_0^{(i)}), \mu_k)$.
  \State Form labeled dataset $\mathcal{D}_{\text{syn}} \gets \{(\boldsymbol{x}_0^{(i)}, y^{(i)})\}$.

  \Statex
  \State \textit{// Step 2: Create conditional model}
  \If{DPM is small (e.g., for CIFAR-10)} 
      \State Train $p_{\phi}^t(y|\boldsymbol{x}_t)$ by minimizing $\mathcal{L}_{\text{cls}}$ on $\mathcal{D}_{\text{syn}}$:
      \State $\min_{\phi} \mathbb{E}_{t,(\boldsymbol{x}_0,y),\boldsymbol\epsilon}\bigl[-\log p_{\phi}^t(y\mid\boldsymbol{x}_t)\bigr]$
  \ElsIf{DPM is large (e.g., Stable Diffusion)} 
      \State Fine-tune LoRA adapters $\Delta\theta$ by minimizing $\mathcal{L}_{\text{LoRA}}$:
      \State $\min_{\Delta\theta} \mathbb{E}_{t, \boldsymbol{x}_0, y \sim \mathcal{D}_{\text{syn}}, \boldsymbol{\epsilon}}[\|\boldsymbol{\epsilon} - \epsilon_{\theta + \Delta\theta}(\boldsymbol{x}_t, t, y)\|^2]$
  \EndIf

  \Statex
  \State \textbf{Part 2: Extract Data with Surrogate Condition}
  \State $\mathcal{D}_{\text{ext}} \gets \emptyset$
  \For{$i=1$ to $N_G$}
      \State Sample target cluster $c\sim\mathcal{U}\{1,\dots,K'\}$
      \State $\boldsymbol{x}_T\sim\mathcal{N}(0,I)$
      \For{$t=T$ \textbf{down to} $1$}
          \If{using classifier guidance}
              \State $s_{\text{guided}} \gets s_{\theta}(\boldsymbol{x}_t,t) + \lambda \cdot \nabla_{\boldsymbol{x}_t}\log p_{\phi}^t(c\mid\boldsymbol{x}_t)$
          \ElsIf{using LoRA fine-tuning}
              \State $s_{\text{guided}} \gets s_{\theta+\Delta\theta}(\boldsymbol{x}_t, t, c)$
          \EndIf
          \State $\boldsymbol{x}_{t-1} \gets \mathrm{DS}(\boldsymbol{x}_t,t,s_{\text{guided}})$
      \EndFor
      \State Append $\boldsymbol{x}_0$ to $\mathcal{D}_{\text{ext}}$
  \EndFor
  \State \Return $\mathcal{D}_{\text{ext}}$
\end{algorithmic}
\end{algorithm}

\section{Theoretical Analysis}

In this section, we first introduce a Kullback-Leibler (KL) divergence-based measure to quantify the degree of memorization in generative models. Building on this, we provide a theoretical explanation for data memorization in conditional DPMs and clarify why \textbf{SIDE} can effectively extract data.

\subsection{Distributional Memorization Measure}

Several approaches exist for measuring the memorization effect in generative models. One common method compares each generated sample to raw training samples individually, for example using $L_p$ distances. While effective for evaluating data extraction performance, such sample-level metrics fall short in assessing the overall memorization behavior of the model. To capture model-level memorization relative to the training data distribution and support our theoretical analysis, we introduce the following distributional memorization measure.

We measure memorization by the KL divergence between the uniform empirical distribution over $\mathcal{D}$, $\frac{1}{|\mathcal{D}|}\sum_{x_i \in \mathcal{D}} \delta(x_i)$ (where $\delta(\cdot)$ is the Dirac delta function), and the distribution $p$ of the model’s generated samples. The $\delta(\cdot)$ function imposes a \textbf{point-wise memorization measure}, quantifying alignment with each original data point. A smaller KL divergence indicates stronger memorization. Since direct computation is infeasible for continuous $p$, we approximate each Dirac delta with a normal distribution of small variance, as shown below.

\begin{Definition}[Memorization Divergence]\label{def:mem}
    Given a generative model $p_{\theta}$ with parameters $\theta$ and training dataset $\mathcal{D}=\{x_i\}_{i=1}^N$, the degree of divergence between $p_{\theta}$ and distribution of training dataset is defined as: 
\begin{equation}
\begin{aligned}\label{eq:1}
\mathcal{M}(\mathcal{D};p_\theta,\epsilon) &= D_{\mathrm{KL}}(q_{\epsilon} \| p_{\theta}) \\
\text{with } q_{\epsilon}(x) = &\frac{1}{N} \sum_{x_i \in \mathcal{D}} \mathcal{N}(x|x_i,\epsilon^2 I),
\end{aligned}
\end{equation}
where \(x_i \in \mathbb{R}^d\) denotes the $i$-th training sample, \(N\) is the total number of training samples, \(p_{\theta}(x)\) represents the probability density function (PDF) of the generated samples, and $\mathcal{N}(x|x_i,\epsilon^2 I)$ is the normal distribution with mean $x_i$ and covariance matrix $\epsilon^2 I$.
\end{Definition}

Note that in Equation~\ref{eq:1}, a smaller value of $\mathcal{M}(\mathcal{D};p_\theta,\epsilon)$ indicates greater overlap between the two distributions, signifying stronger memorization. As $\epsilon$ approaches $0$, the measured memorization divergence becomes more precise. In fact, the normal distribution $\mathcal{N}(x|x_i,\epsilon^2 I)$ can be replaced with any continuous distribution family $\hat{q}(x|x_j,\epsilon)$ that (1) is symmetric with respect to $x$ and $x_j$ (i.e., $\hat{q}(x|x_j,\epsilon) = \hat{q}(x_j|x,\epsilon)$), and (2) converges to $\delta(x_i)$ in distribution. This substitution does not affect \cref{theorem_informative}.

While one might be concerned about the effect of $\epsilon$ on the divergence, this measure is primarily intended for comparative analysis when $\epsilon$ is sufficiently small. \textbf{When comparing memorization divergence across different models, $\epsilon$ does not affect the results, as demonstrated in the Appendix.}

\subsection{Theoretical Analysis}\label{sec:3.2}
Building on the memorization divergence measure, we provide a theoretical analysis to explain why conditional DPMs exhibit a stronger memorization effect. Our analysis focuses on the concept of \emph{informative labels}, which partition a dataset into multiple disjoint subsets. We show that DPMs conditioned on informative labels tend to demonstrate enhanced memorization.

\paragraph{Informative Labels}
\label{subsection:informative_labels}

The concept of \emph{informative labels} has previously been discussed in the context of class labels~\cite{Gu2023OnMI}. In this work, we generalize this notion to include both class labels and random labels as special cases. Formally, we define an informative label as follows:

\begin{Definition}[Informative Label]\label{def:Informative_label}
Let $\mathcal{Y}$ be a data attribute taking values in $\{y_i\}_{i =1}^{C}$.
We define $\mathcal{Y}$ as an informative label if it enables the partitioning of the dataset into mutually disjoint subsets $ \{ \mathcal{D}_i \}_{i = 1}^{C}$, where each subset corresponds to a distinct value of $\mathcal{Y}$.
\end{Definition}

In this definition, informative labels are not limited to traditional class labels; they can also include text captions, features, or cluster information that group training samples into subsets. The key requirement is that an informative label must distinguish one subset of samples from others. An extreme case is when all samples share the same label, making it non-informative. By this definition, both class-wise and random labels are special cases of informative labels. Informative labels may be explicit—such as class labels, random labels, or text captions—or implicit, such as salient clusters.

Next, we present our main theoretical result on the memorization mechanism of conditional DPMs and provide insight into why \textbf{SIDE} is effective.
Let $\mathcal{D}_i$ represent the subset of data with informative label $\mathcal{Y} = y_i$. We denote the overall data distribution of the original dataset $\mathcal{D}$ by $p$, and the corresponding subset distribution by $p_i$ for each attribute $y_i$.

\begin{theorem}
\label{theorem_informative}
If a generative model \( p_{\theta_{i}} \) matches the target distribution \( p_i \) almost everywhere for the informative label \( y_i \), that is, \( TV(p_i, p_{\theta_{i}}) = 0 \), then with probability 1:
\begin{align}
\lim_{\epsilon \to 0}\lim_{|\mathcal{D}_i| \to \infty}\left(\mathcal{M}(\mathcal{D}_i; p_{\theta_i}, \epsilon) - \mathcal{M}(\mathcal{D}_i; p_{\theta}, \epsilon)\right) \le 0,
\end{align}
where $TV(\cdot)$ denotes the total variance distance, and \( p_{\theta_{i}} \) and $p_{\theta}$ denote the distribution of generated data for model trained on data labeled $y_i$ and on the entire dataset, respectively. Equality holds if and only if \( TV(p, p_{i}) = 0 \).
\end{theorem}

The proof for Theorem \ref{theorem_informative} is provided in Appendix \ref{proof:theorem2}. This theorem shows that conditioning on informative labels enhances memorization. While any form of conditioning can help, its effectiveness depends on how well it isolates a specific, high-density region of the data distribution. Conventional text prompts or class labels offer only coarse guidance by pointing to broad concepts. In contrast, SIDE delivers fine-grained guidance by first identifying the DPM’s native data clusters, which are dense groups of similar images formed internally by the model, and then targeting these clusters. This approach aligns the extraction attack with the model’s intrinsic data representation.

\begin{table*}
\centering

\begin{tabular}{l|c|cccccccc}
\toprule
\multirow{2}{*}{Dataset} & \multirow{2}{*}{Method} & \multicolumn{2}{c}{Low Similarity} & \multicolumn{2}{c}{Mid Similarity} & \multicolumn{2}{c}{High Similarity} & \multirow{2}{*}{\begin{tabular}[c]{@{}c@{}}95th SSCD \\ Percentile\end{tabular}} & \multirow{2}{*}{\begin{tabular}[c]{@{}c@{}}95th  \\ $L_2$ Dist.\end{tabular}} \\
& & \small AMS(\%) & \small UMS(\%) & \small AMS(\%) & \small UMS(\%) & \small AMS(\%) & \small UMS(\%) & & \\
\midrule

\multirow{3}{*}{\small CIFAR-10}
& \small Carlini UnCond &           2.470 &           1.770 &           0.910 &           0.710 &           0.510 &           0.420 & / & 1.85 \\
& \small Carlini Cond &           5.250 &          2.020 &           2.300 &           0.880 &           1.620 &           0.640 & / & 1.62 \\
& \small \textbf{SIDE (Ours)}  &  \textbf{7.830} &  \textbf{2.730} &  \textbf{3.830} &  \textbf{1.190} &  \textbf{2.610} &  \textbf{0.760} & \textbf{/} & \textbf{1.41} \\
\midrule 
\multirow{3}{*}{\small CelebA-HQ-FI}
& \small Carlini UnCond & 11.656 & 2.120 & 0.596 & 0.328 & 0.044 & 0.040 & 0.433 & / \\
& \small Carlini Cond  & 15.010 & 2.624 & 1.310 & 0.554 & 0.090 & 0.082 & 0.485 & / \\
& \small \textbf{SIDE (Ours)} & \textbf{23.266} & \textbf{4.198} & \textbf{2.227} & \textbf{0.842} & \textbf{0.141} & \textbf{0.148} & \textbf{0.543} & \textbf{/} \\

\midrule

\multirow{3}{*}{\small CelebA-25000}
& \small Carlini UnCond & 5.000 & 4.240 & 0.100 & 0.100 & 0.000 & 0.000 & 0.404 & / \\
& \small Carlini Cond & 8.712 & 6.802 & 0.234 & 0.234 & 0.010 & 0.010 & 0.439 & / \\
& \small \textbf{SIDE (Ours)} & \textbf{20.527} & \textbf{11.446} & \textbf{1.842 }& \textbf{1.164} & \textbf{0.030} & \textbf{0.030} & \textbf{0.542} & \textbf{/} \\

\midrule 
\multirow{3}{*}{\small CelebA}
& \small Carlini UnCond &           1.953 &           1.895 &            0.000 &           0.000 &           0.000 &           0.000 & 0.404 & / \\
& \small Carlini Cond  &  4.682 & 4.706 & 0.098 & 0.098 & 0.000 & 0.000 & 0.436 & / \\
& \small \textbf{SIDE (Ours)} & \textbf{7.187} & \textbf{6.582} & \textbf{0.273 }& \textbf{0.273} & \textbf{0.023} & \textbf{0.023} & \textbf{0.501} & \textbf{/} \\

\midrule 
\multirow{3}{*}{\small ImageNet}
& \small Carlini UnCond &           0.000 &           0.000 &           0.000 &           0.000 &           0.000 &           0.000 & 0.250 & / \\
& \small Carlini Cond &   0.152 &  0.152 &  0.076 &  0.076 & 0.000&  0.000  & 0.283 & / \\
& \small \textbf{SIDE (Ours)} & \textbf{0.443} & \textbf{0.239} & \textbf{0.231} & \textbf{0.231} & \textbf{0.039} & \textbf{0.039} & \textbf{0.347} & \textbf{/} \\

\midrule 
\multirow{3}{*}{\small LAION-5B}
& \small Carlini UnCond &           0.000 &           0.000 &           0.000 &           0.000 &           0.000 &           0.000 & 0.215 & / \\
& \small Carlini Cond & 0.371 & 0.006 & 0.247 & 0.004 & 0.096 & 0.003 & 0.253 & / \\
& \small \textbf{SIDE (Ours)} & \textbf{2.221} & \textbf{0.013} & \textbf{0.805} & \textbf{0.007} & \textbf{0.131} & \textbf{0.006} & \textbf{0.394} & \textbf{/} \\
\bottomrule

\end{tabular}
% \caption{Comprehensive performance comparison of our SIDE method against baseline unconditional (UnCond) and conditional (Cond) extraction attacks from Carlini et al.~\cite{carlini2023extracting} across multiple datasets. SIDE consistently demonstrates superior performance, achieving significantly higher AMS and UMS percentages. Furthermore, it yields better 95th percentile similarity scores (higher SSCD for high-resolution datasets, lower L2 distance for CIFAR-10), indicating both a higher rate and higher quality of extractions.}
\caption{Performance comparison of our SIDE method with baseline unconditional (Carlini UnCond) and conditional (Carlini Cond) extraction attacks from~\cite{carlini2023extracting} across multiple datasets. %  \vspace{-1.5em}
% SIDE consistently outperforms the baselines, achieving substantially higher AMS, UMS percentages, and 95th percentile similarity scores.
}
\label{celeba_25000_table}

\end{table*}

\begin{figure}[h]
    \centering
    \includegraphics[width=0.45\textwidth]{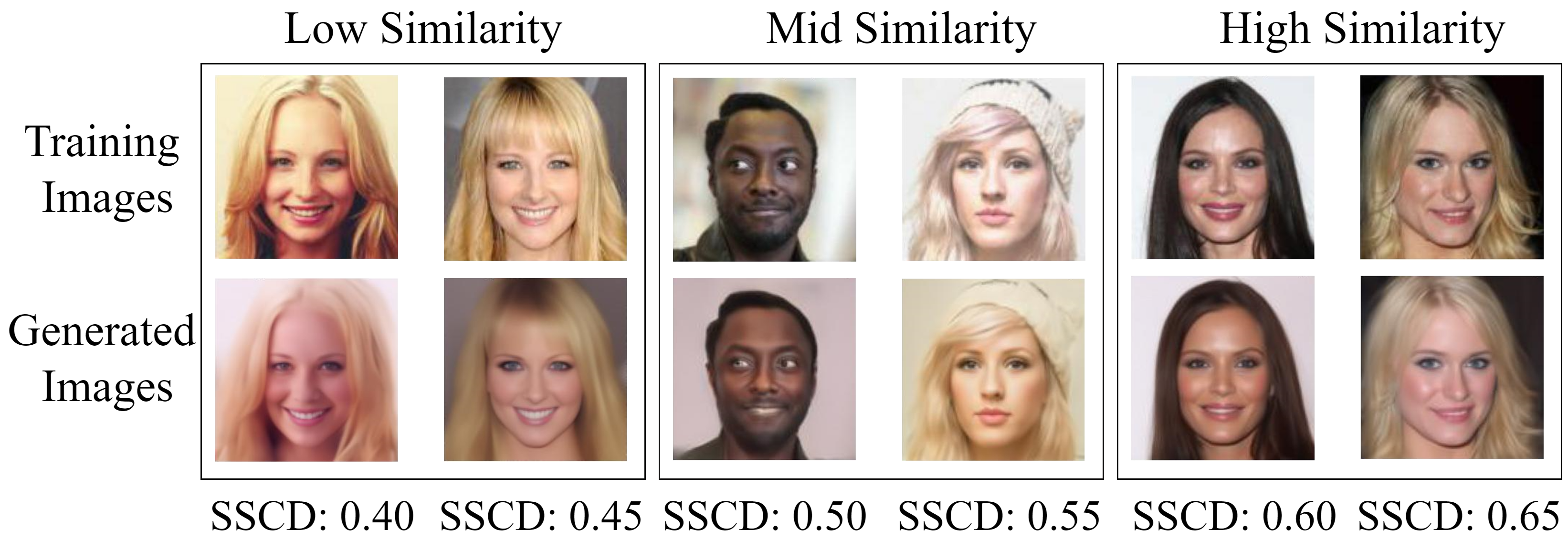}

    \caption{A comparison between original training images (top row) and images extracted by our SIDE method (bottom row). The matched pairs are categorized by similarity: low (SSCD score $<$ 0.5), mid (SSCD score between 0.5 and 0.6), and high (SSCD score $>$ 0.6), illustrating the varying degrees of semantic resemblance achieved by SIDE.}
        \label{fig:comparison}
\end{figure}
\section{Experiments}
In this section, we first present the performance metrics and experimental setup, followed by the main evaluation results. We also include an ablation study and hyperparameter analysis to provide deeper insight into the mechanisms of SIDE.

\subsection{Image-level Performance Metrics}
Determining whether an extracted image is a memorized copy of a training sample is challenging. Pixel-space distances such as $L_p$ are ineffective for semantically similar but non-identical images. Prior work~\cite{Somepalli2022DiffusionAO,Gu2023OnMI} uses the 95th percentile Self-Supervised Descriptor for Image Copy Detection (SSCD) score, but this approach has notable limitations: (1) it fails to measure the uniqueness of memorized content; (2) it can underestimate the total number of memorized samples; and (3) it does not account for different types of memorization.

To address these issues, we propose two new metrics: \textbf{Average Memorization Score (AMS)} and \textbf{Unique Memorization Score (UMS)}.
\begin{align}
& \text{AMS}\left( {\mathcal{D}}_{1},\mathcal{D}_{2},\alpha ,\beta \right) =\frac{\sum_{x_i\in \mathcal{D}_{1}}^{}{\mathcal{F} \left(  x_i,\mathcal{D}_{2} ,\alpha, \beta \right)}}{N_G} \\
& \text{UMS}\left( {\mathcal{D}}_{1},\mathcal{D}_{2},\alpha ,\beta \right) = \frac{|\bigcup_{x_i\in \mathcal{D}_{1}} \phi \left( x_{i},\mathcal{D}_{2} ,\alpha, \beta \right) |}{N_G},
\end{align}
where ${\mathcal{D}}_1$ is the set of $N_G$ generated images and $\mathcal{D}_2$ is the training set. These metrics rely on helper functions that check whether the similarity $\gamma(x_i, x_j)$ between a generated image $x_i$ and any training image $x_j$ falls within a threshold range $[\alpha, \beta]$:
\begin{align}
    & \mathcal{F}(x_i, \mathcal{D}_{2}, \alpha, \beta) = \mathds{1} \Big[\max_{x_j \in \mathcal{D}_{2}} \gamma(x_i, x_j) \in [\alpha,\beta] \Big] \\
    & \phi(x_i, \mathcal{D}_{2}, \alpha, \beta) = \{j : x_j \in \mathcal{D}_{2}, \gamma(x_i, x_j) \in [\alpha,\beta] \}
\end{align}
For the similarity function $\gamma$, we use the normalized $L_2$ distance for low-resolution datasets \cite{carlini2023extracting} and the SSCD score for high-resolution datasets.

We categorize memorization into \textbf{low}, \textbf{mid}, and \textbf{high} similarity levels by applying different $[\alpha, \beta]$ thresholds. This enables a more granular assessment of memorization—from near-exact copies to broader stylistic influence—which is especially important for copyright analysis~\cite{lee2023talkin,sag2023copyright,sobel2023elements}.

\paragraph{Relation to Existing Metrics.}
While similar metrics have been proposed~\cite{carlini2023extracting,chen2024towards}, ours are the first to explicitly incorporate varying similarity levels. Additionally, our UMS uniquely accounts for the number of generated images $N_G$, a factor overlooked in~\cite{carlini2023extracting}. The effect of $N_G$ is non-linear, as captured by the expected number of unique memorized samples:
$ \mathbb{E}[N_{\mathrm{umem}}] = \sum_{i=1}^M{1-\left( 1-p_{\gamma}\left( i \right) \right) ^{N_G}} $
This underscores the importance of comparing UMS scores under a constant $N_G$. Lastly, note that AMS and UMS are individual-level metrics, distinct from distributional measures such as the one defined in Equation~\ref{def:mem}.

\subsection{Experimental Setup}

We evaluated our method on 6 datasets: CIFAR-10, three CelebA variants (CelebA-HQ-FI~\cite{na2022unrestricted}, CelebA-25000, and full CelebA~\cite{liu2015faceattributes}, all 128$\times$128), ImageNet~\cite{deng2009imagenet} (256$\times$256), and LAION-5B (512$\times$512)\cite{schuhmann2022laionb} using a pre-trained Stable Diffusion 1.5 model. For models trained from scratch, we used a DDIM scheduler\cite{song2021denoising} from the HuggingFace implementation~\cite{diffusers} with a batch size of 64. Training was run for approximately 2048 epochs on CIFAR-10, 3000 on CelebA-HQ-FI, 1000 on the other CelebA sets, and 1980K steps on ImageNet, which was evaluated on the ImageNette subset~\cite{howard2020fastai}. All images were normalized to $[-1, 1]$.
For surrogate guidance, we used a ResNet34 pseudo-labeler~\cite{He2015DeepRL}, an SSCD feature extractor with 100 clusters, and a cohesion threshold of 0.5. LoRA fine-tuning for Stable Diffusion used a rank of 512. The time-dependent classifier was trained with AdamW~\cite{loshchilov2018decoupled} at a learning rate of 1e-4, and LoRA fine-tuning at 1e-5. On LAION-5B, we evaluated extraction against known memorized images~\cite{hong2024membenchmemorizedimagetrigger}.

% \begin{table}
% \centering
% \caption{The coefficient and $R^2$ results of a fitted linear model between the number of classes vs. AMS/UMS. $R^2$ measures how well a regression model fits the data (1 is perfect fitting). }

% \label{tab:relationship}
% \begin{tabular}{lcccc}
% \hline
% \textbf{Relationship} & \textbf{Coefficient} & \textbf{$R^2$} \\
% \hline
% Class vs. AMS & 7.4 $\times 10^{-5}$ & 0.637 \\
% Class vs. UMS & 6.1 $\times 10^{-5}$ & 0.483 \\
% \hline
% \end{tabular}

% \end{table}

\subsection{Main Results}
\label{subsec:main_results}
% We compare SIDE with the \emph{Random} baseline proposed by Carlini et al.\cite{carlini2023extracting} and a variant of SIDE that substitutes the time-dependent classifier with a standard (time-independent) classifier. Here, "TD" refers to the time-dependent classifier trained using our proposed TDKD method, "TI" denotes the time-independent classifier, and "OL" indicates training with original dataset labels. 

% The \emph{Random} baseline generates images directly using the target unconditional DPM, as described in \cite{carlini2023extracting}. We average the results across various values of \(\lambda\) (defined in \eqref{classifier_sampling}), ranging from 5 to 10, with a detailed analysis provided in Section \ref{hyper_analysis}. It is important to note that \(\lambda = 0\) corresponds to the \emph{Random} baseline.
We evaluate our SIDE method against two state-of-the-art baselines introduced by \citet{carlini2023extracting}: \textbf{Carlini UnCond}, which samples unconditionally from the target model, and \textbf{Carlini Cond}, which uses a standard, time-independent classifier for conditional guidance. As noted in~\cite{Fang2024PrivacyLO,Ma2025SafetyAS}, these remain the only established methods for extracting training data from pretrained DPMs, making them the most relevant benchmarks for assessing the effectiveness of SIDE’s surrogate guidance mechanism. For evaluation, we generate 51,200 images for CelebA-HQ-FI, 50,000 for CelebA-25000, 10,000 for CIFAR-10, 5,120 for CelebA, 2,560 for ImageNet, and 512,000 for LAION-5B. The results are reported in \cref{celeba_25000_table}.
% This effort results in one of the most enormous generated image datasets for studying the memorization of DPMs.

% \begin{figure*}
%     \centering
%     \includegraphics[width=0.95\textwidth]{figures/celeba-5000_new.png}

%     \caption{Hyper-parameter ($\lambda$) analysis on CelebA-HQ-FI. For high similarity, the best $\lambda$ for AMS and UMS are 16 and 13. For other similarity levels, the best $\lambda$ for AMS and UMS is 13.}
%     \label{fig:celeba_5000}
% \end{figure*}

% \begin{figure}[ht]
%     \centering
%     \includegraphics[width=0.4\textwidth]{figures/impact_on_classifier_choice.png}
%     \caption{The relationship between the number of classes and AMS/UMS. The fitted regression lines demonstrate a positive correlation in both metrics.}
%     \label{fig:enter-label}
% \end{figure}

\paragraph{Effectiveness of SIDE.} 

% \cref{celeba_25000_table} summarizes the AMS and UMS results for the five datasets (results of 95th percentile SSCD are in Appendix \ref{more_experiments_ams_ums}). SIDE shows strong performance in extracting training data across various datasets and similarity levels.  Data extraction of unconditional DPM on large-scale datasets was considered impossible. But \textbf{SIDE} successfully extracts from these datasets.

% On CIFAR-10, SIDE outperforms baselines at all similarity levels. For low similarity, it achieves 5.320\% AMS (vs. Random's 2.470\%) and 2.05\% UMS (vs. 1.780\%). For mid similarity, SIDE reaches 2.490\% AMS and 0.860\% UMS, surpassing Random's 0.910\% AMS and 0.710\% UMS. At high similarity, SIDE achieves 1.770\% AMS and 0.560\% UMS, outperforming Random's 0.510\% AMS and 0.420\% UMS.

% On CelebA-HQ-FI, SIDE boosts mid-level AMS by 87\% to 1.115\% and UMS by 37\% to 0.444\%, with a 20\% average improvement across other levels. SIDE could extract 900 high, 11,500 mid, and 151,720 low similarity images from 1 million samples. On CelebA-25000, SIDE improves AMS and UMS by 75\% and 63\% for low similarity, and 124\% and 112\% for mid similarity, excelling at high-similarity memorized data extraction. For CelebA and ImageNet, SIDE successfully extracts training data where baselines fail.

The results in Table~\ref{celeba_25000_table} clearly demonstrate the effectiveness of our SIDE method, which consistently and significantly outperforms both unconditional and conditional baselines across all six datasets. For our primary metrics, AMS and UMS, SIDE achieves the highest scores at every similarity level (low, mid, and high) indicating that it extracts not only more memorized samples, but also a greater diversity of unique instances. For example, on CelebA-25000, SIDE achieves a low-similarity AMS of 20.527\%, more than double the 8.712\% of the next best method, Carlini Cond. This trend holds for standard metrics as well: SIDE attains the highest 95th percentile SSCD scores on all high-resolution datasets and the lowest (best) 95th percentile $L_2$ distance on CIFAR-10. The consistent superiority of SIDE across diverse datasets and multiple evaluation metrics validates the effectiveness of our surrogate guidance approach. Notably, SIDE can even surpass the extraction performance of conditional DPMs when applied to unconditional DPMs.

\begin{figure}[ht]
    \centering
    \includegraphics[width=0.8\linewidth]{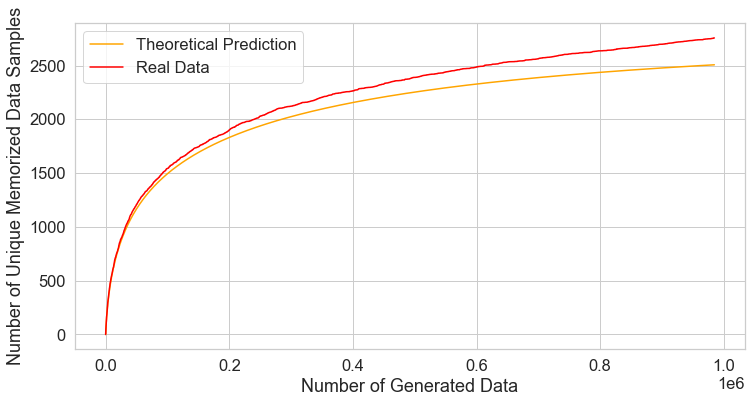}
    
    \caption{Validation of \(N_G\) 's significance. }% \vspace{-1.5em}
    
    \label{fig:ums_val}
    
\end{figure}
% \paragraph{Impact of the Classifier on SIDE}
% The classifier used to train SIDE is associated with a certain number of classes. Here, we explore the relationship between the number of classes (of the classifier) and the extraction performance at a low similarity level, using 1,200 images per class with \(\lambda = 5\) on the CelebA-HQ-FI dataset.
% As shown in \cref{tab:relationship} and \cref{fig:enter-label}, the number of classes exhibits a strong positive correlation with both AMS and UMS. Increasing the number of classes enhances both metrics, with a more pronounced effect on AMS. The UMS values here differ significantly from those in \cref{celeba_25000_table} due to the smaller sample size per class (1,200 images) in this experiment. The stronger improvement in AMS arises because the classifier captures fine-grained details more effectively with more classes, leading to more accurate matches at lower similarity levels. In contrast, UMS is less affected, as it depends more on overall image diversity, which is constrained by the smaller sample size per class in this setting.

\paragraph{Importance of $N_G$ in UMS.} The number of uniquely memorized samples in a dataset of size $M$ can be formulated as $\sum_{i=1}^M{1-\left( 1-k_i \right)^{N_G}}$, where $k_i$ denotes the probability the $i$-th sample is extracted per trial.
To empirically verify the importance of $N_G$ , we generate 1 million samples using a DPM trained on CelebA-HQ-FI. As shown in \cref{fig:ums_val}, the theoretical and empirical results align closely, confirming that $N_G$ non-linearly influences UMS.

% Unlike \cite{carlini2023extracting} that overlooked $N_G$'s significance, UMS, supported by both our theoretical derivation and experimental validation in \cref{fig:ums_val}, enables more rigorous and fair evaluations.

% \paragraph{Hyper-parameter Analysis}
% \label{hyper_analysis}

% Here, we test the sensitivity of SIDE to its hyper-parameter $\lambda$. We generate 50,000 images for each integer value of $\lambda$ within the range of [0, 50]. As shown in Figure \ref{fig:celeba_5000}, the memorization score increases at first, reaching its highest, then decreases as $\lambda$ increases. This can be understood from sampling SDE \eqref{classifier_sampling}. Starting from 0, the diffusion models are unconditional. As $\lambda$ increases, the diffusion models become conditional, and according to \cref{theorem_informative}, the memorization effect will be triggered. However, when $\lambda$ becomes excessively large, the generated images will overfit the classifier's decision boundaries, resulting in reduced diversity and a failure to accurately reflect the underlying data distribution. Consequently, the memorization score decreases.
\begin{figure}[ht]
    \centering
    \includegraphics[width=0.8\linewidth]{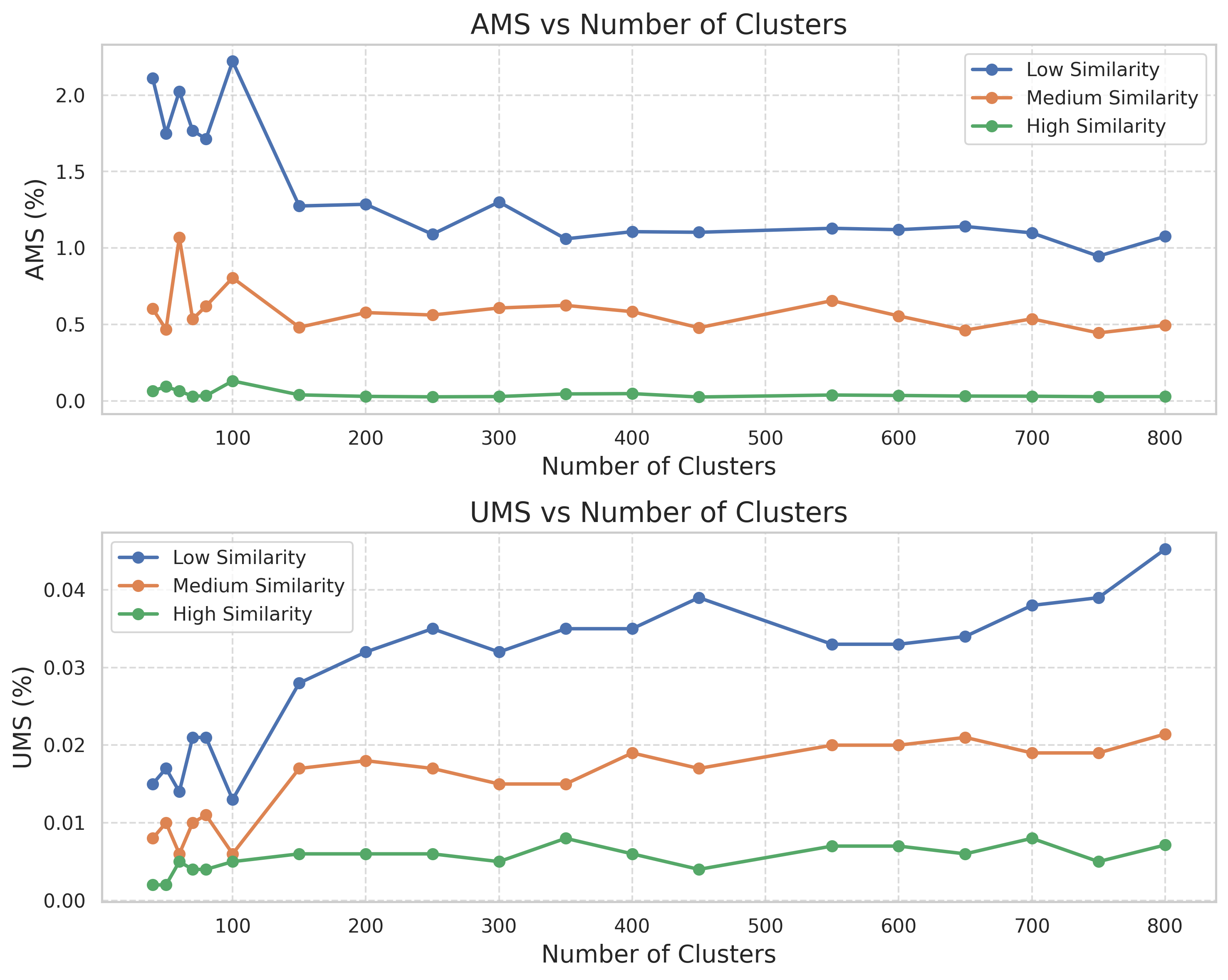}
    
    \caption{Effects of clusters number ($K$) on AMS and UMS.}%\vspace{-0.5em} }
    \label{fig:combined_ams_ums_plot}
    
\end{figure}
\paragraph{Influence of the Number of Clusters.}
We analyze how the number of clusters, $K$, affects extraction performance, as shown in \cref{fig:combined_ams_ums_plot} on the LAION-5B dataset. The results reveal a clear trade-off. With fewer clusters ($K < 200$), AMS is volatile, suggesting that a moderate $K$ is optimal for AMS. In contrast, as $K$ increases ($K > 400$), UMS steadily rises while AMS shows a slight decline. This suggests that a larger $K$ creates more specific, high-purity clusters that enhance the diversity of unique extractions, even if the overall likelihood of a match decreases. Thus, the optimal value of $K$ depends on the attack objective: whether the priority is maximizing hit rate or extraction diversity.

\begin{figure}[ht]
    \centering
    \includegraphics[width=1.0\linewidth]{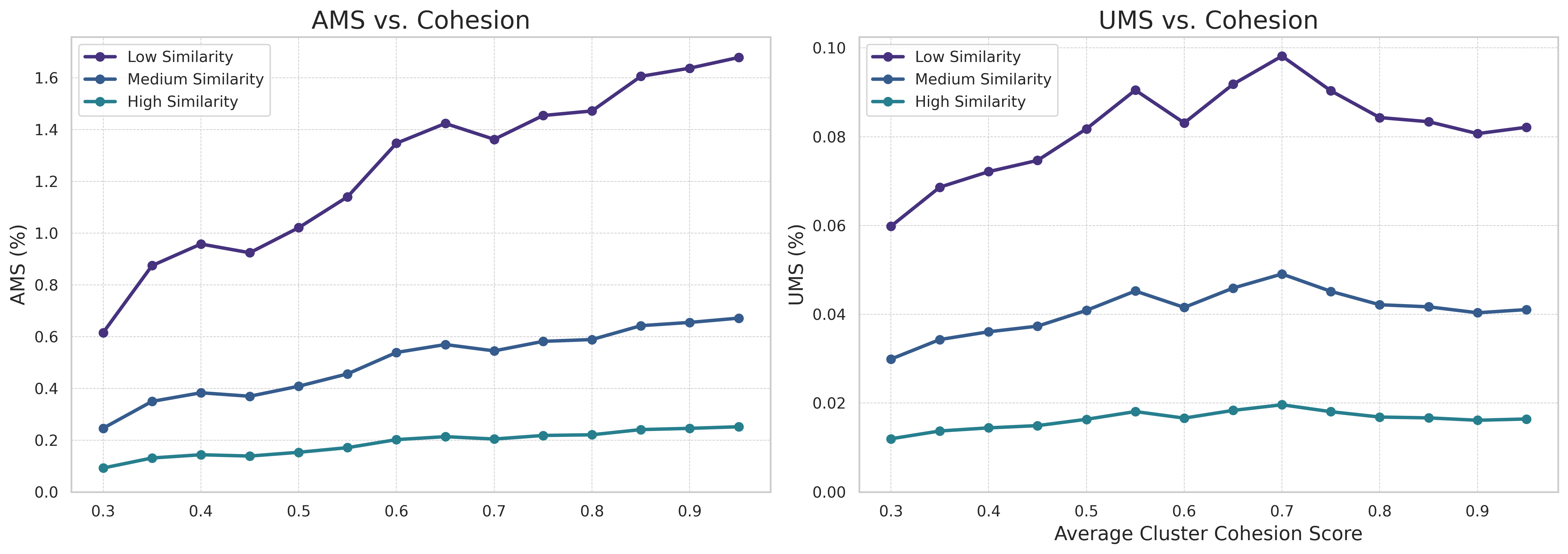}
   
    \caption{The impact of cluster cohesion on AMS and UMS.}%\vspace{-0.5em} }
    \label{fig:cohesion}
    
\end{figure}

\paragraph{Analysis of Cluster Cohesion.}
We examine the effect of cluster cohesion on extraction performance using LAION-5B, averaging results over 50, 100, 150, and 200 clusters, as shown in \cref{fig:cohesion}. The results reveal a critical trade-off: AMS increases consistently with higher cohesion, while UMS peaks at a cohesion score of around 0.6 before declining. This occurs because increasing cohesion improves the ability of surrogate labels to isolate uniquely memorized samples, but beyond this peak, clusters become overspecialized. As a result, AMS improves, while UMS suffers. 

\begin{figure}[h]
    \centering
    % Make sure this filename matches the image you save
    \includegraphics[width=1\linewidth]{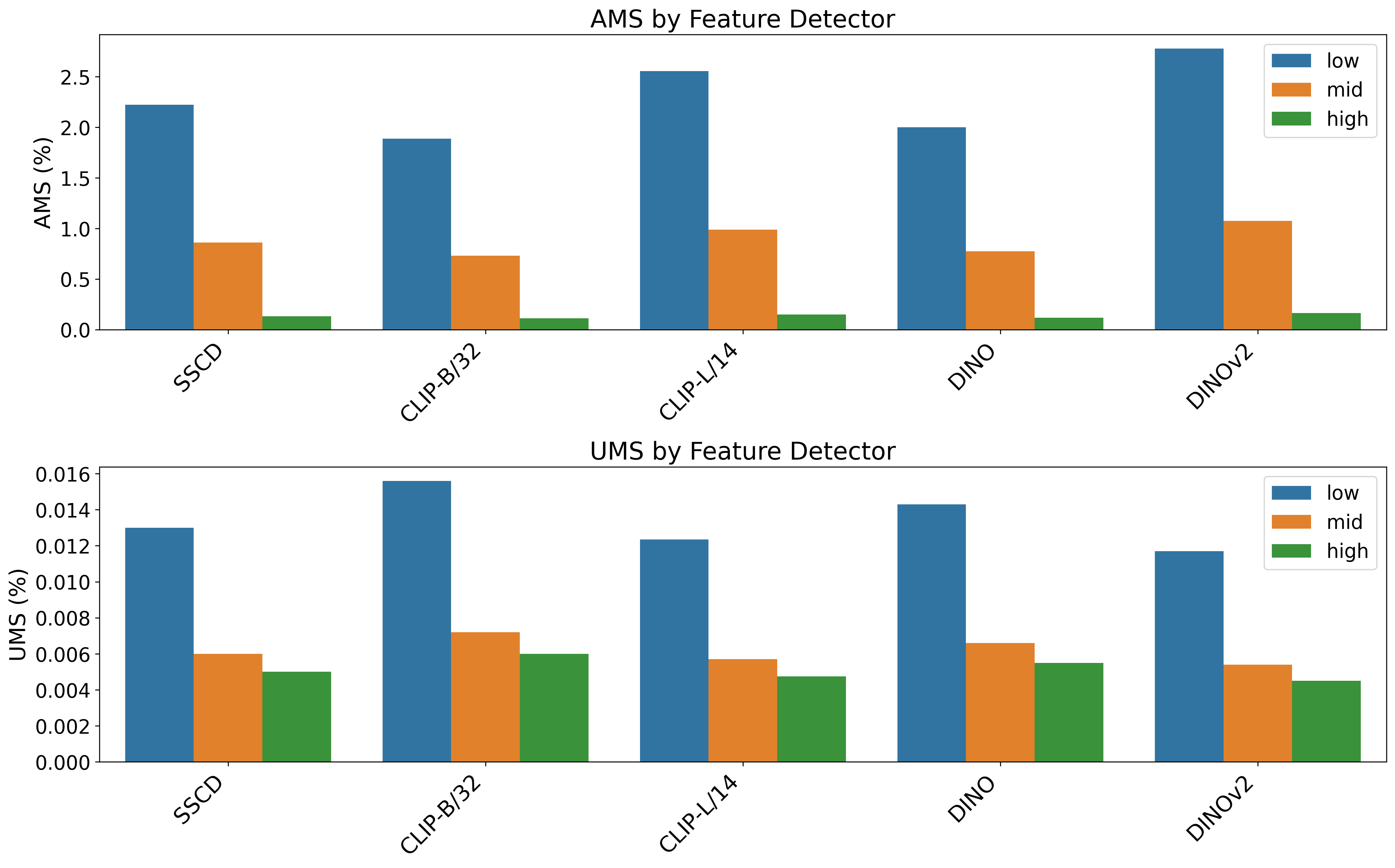}
    \caption{Effects of feature extractor on AMS and UMS.}% \vspace{-0.5em}}
    \label{fig:feature_extractor_ablation}
\end{figure}

\paragraph{Robustness to Feature Extractor Choice}
To assess robustness, we evaluated SIDE's performance using various state-of-the-art feature extractors (e.g., CLIP, DINOv2, SSCD) to generate surrogate labels. As shown in Figure \ref{fig:feature_extractor_ablation}, while minor variations exist, the choice of extractor does not significantly impact the attack's success. All tested models yielded consistently high AMS and UMS scores, confirming that SIDE is a robust and broadly applicable framework, not reliant on a single feature extractor. \textbf{For additional hyperparameter analysis, please refer to the Appendix.}

% \section{Limitations:}  SIDE has certain limitations that warrant further research. SIDE mainly uses human-labelled data to extract data, which is scarce and limits its practical use. Future work could investigate how to generating implicit labels.
% , such as employing unsupervised clustering or utilizing pretrained models to label synthetic datasets
% Future work could explore hybrid methods for generating implicit labels by using unsupervised clustering or leveraging pre-trained models to generate labels for synthetic datasets. 

% \paragraph{Limitation.}
% The scope of this study is focused on image diffusion models, lacking other modalities, such as text generation with Large Language Models (LLMs). This remains a direction for future research.

\section{Conclusion}
In this work, we introduced SIDE, a novel data extraction framework that exploits memorization in diffusion probabilistic models (DPMs) by constructing precise surrogate conditions. Supported by a theoretical analysis of informative labels, our experiments demonstrated that SIDE consistently outperformed existing baselines.
Notably, SIDE successfully extracted data from unconditional DPMs, which were previously considered safe, and achieved effectiveness that surpassed attacks on explicitly conditional models.
These findings highlight precise conditioning as a critical vector for data leakage and establish SIDE as a new benchmark for developing and evaluating defenses against data extraction in generative models.

% In this paper, we proposed a novel method \emph{SIDE} for extracting training data from unconditional diffusion probabilistic models (DPMs) by constructing surrogate conditions. At its core, SIDE trains a time-dependent classifier to create these surrogate conditions. We empirically validated the effectiveness of SIDE on datasets of varying scales, including CelebA, ImageNet, and CIFAR-10. Additionally, we provide a theoretical analysis to explain the role of information labels in facilitating memorization in DPMs.
% Our work highlights the potential for memorization in unconditional DPMs and demonstrates that such memorized images can be extracted by constructing surrogate conditions. Our study advances the understanding of memorization mechanisms in DPMs and motivates the development of more effective memorization mitigation strategies, such as regularizing deep representations to be less informative.


%%%%%%%%% REFERENCES

{
\bibliographystyle{ieee_fullname}
\bibliography{aaai2026}

\begin{thebibliography}{10}\itemsep=-1pt

\bibitem{achilli2025memorization}
Beatrice Achilli, Luca Ambrogioni, Carlo Lucibello, Marc M{\'e}zard, and Enrico Ventura.
\newblock Memorization and generalization in generative diffusion under the manifold hypothesis.
\newblock {\em arXiv preprint arXiv:2502.09578}, 2025.

\bibitem{asay2020independent}
Clark~D Asay.
\newblock Independent creation in a world of ai.
\newblock {\em FIU Law Review}, 14:201, 2020.

\bibitem{baptista2025memorization}
Ricardo Baptista, Agnimitra Dasgupta, Nikola~B Kovachki, Assad Oberai, and Andrew~M Stuart.
\newblock Memorization and regularization in generative diffusion models.
\newblock {\em arXiv preprint arXiv:2501.15785}, 2025.

\bibitem{BetkerImprovingIG}
James Betker, Gabriel Goh, Li Jing, TimBrooks, Jianfeng Wang, Linjie Li, LongOuyang, JuntangZhuang, JoyceLee, YufeiGuo, WesamManassra, PrafullaDhariwal, CaseyChu, YunxinJiao, and Aditya Ramesh.
\newblock Improving image generation with better captions.
\newblock 2023.

\bibitem{brokman2025identifying}
Jonathan Brokman, Amit Giloni, Omer Hofman, Roman Vainshtein, Hisashi Kojima, and Guy Gilboa.
\newblock Identifying memorization of diffusion models through p-laplace analysis.
\newblock In {\em International Conference on Scale Space and Variational Methods in Computer Vision}, pages 295--307. Springer, 2025.

\bibitem{videoworldsimulators2024}
Tim Brooks, Bill Peebles, Connor Holmes, Will DePue, Yufei Guo, Li Jing, David Schnurr, Joe Taylor, Troy Luhman, Eric Luhman, Clarence Ng, Ricky Wang, and Aditya Ramesh.
\newblock Video generation models as world simulators.
\newblock 2024.

\bibitem{butterick2023stable}
Matthew Butterick.
\newblock Stable diffusion litigation{\textperiodcentered} joseph saveri law firm \& matthew butterick.
\newblock 2023.

\bibitem{carlini2023extracting}
Nicolas Carlini, Jamie Hayes, Milad Nasr, Matthew Jagielski, Vikash Sehwag, Florian Tramer, Borja Balle, Daphne Ippolito, and Eric Wallace.
\newblock Extracting training data from diffusion models.
\newblock In {\em USENIX Security 2023}, 2023.

\bibitem{carlini2022quantifying}
Nicholas Carlini, Daphne Ippolito, Matthew Jagielski, Katherine Lee, Florian Tramer, and Chiyuan Zhang.
\newblock Quantifying memorization across neural language models.
\newblock {\em arXiv preprint arXiv:2202.07646}, 2022.

\bibitem{chen2025enhancing}
Chen Chen, Daochang Liu, Mubarak Shah, and Chang Xu.
\newblock Enhancing privacy-utility trade-offs to mitigate memorization in diffusion models.
\newblock In {\em CVPR 2025}, pages 8182--8191, 2025.

\bibitem{chen2024towards}
Chen Chen, Daochang Liu, and Chang Xu.
\newblock Towards memorization-free diffusion models.
\newblock In {\em CVPR 2024}, 2024.

\bibitem{chen2023trojdiff}
Weixin Chen, Dawn Song, and Bo Li.
\newblock Trojdiff: Trojan attacks on diffusion models with diverse targets.
\newblock In {\em CVPR 2023}, pages 4035--4044, 2023.

\bibitem{chen2024deconstructing}
Xinlei Chen, Zhuang Liu, Saining Xie, and Kaiming He.
\newblock Deconstructing denoising diffusion models for self-supervised learning.
\newblock {\em arXiv preprint arXiv:2401.14404}, 2024.

\bibitem{chen2024comprehensive}
Yunhao Chen, Zihui Yan, and Yunjie Zhu.
\newblock A comprehensive survey for generative data augmentation.
\newblock {\em Neurocomputing}, 2024.

\bibitem{chen2023data}
Yunhao Chen, Zihui Yan, Yunjie Zhu, Zhen Ren, Jianlu Shen, and Yifan Huang.
\newblock Data augmentation for environmental sound classification using diffusion probabilistic model with top-k selection discriminator.
\newblock In {\em ICIC 2023}, 2023.

\bibitem{cooper2024files}
A.~Feder Cooper and James Grimmelmann.
\newblock The files are in the computer: Copyright, memorization, and generative ai.
\newblock {\em arXiv preprint arXiv:2404.12590}, 2024.

\bibitem{dar2024unconditional}
Salman Ul~Hassan Dar, Marvin Seyfarth, Isabelle Ayx, Theano Papavassiliu, Stefan~O Schoenberg, Robert~Malte Siepmann, Fabian~Christopher Laqua, Jannik Kahmann, Norbert Frey, Bettina Bae{\ss}ler, et~al.
\newblock Unconditional latent diffusion models memorize patient imaging data: Implications for openly sharing synthetic data.
\newblock {\em arXiv preprint arXiv:2402.01054}, 2024.

\bibitem{deng2009imagenet}
Jia Deng, Wei Dong, Richard Socher, Li-Jia Li, Kai Li, and Li Fei-Fei.
\newblock Imagenet: A large-scale hierarchical image database.
\newblock In {\em CVPR 2009}, 2009.

\bibitem{dhanuka2025magic}
Gunjan Dhanuka, Sumukh~K Aithal, Avi Schwarzschild, Zhili Feng, J~Zico Kolter, Zachary~Chase Lipton, and Pratyush Maini.
\newblock {MAGIC}: Diffusion model memorization auditing via generative image compression.
\newblock In {\em The Impact of Memorization on Trustworthy Foundation Models: ICML 2025 Workshop}, 2025.

\bibitem{dhariwal2021diffusion}
Prafulla Dhariwal and Alexander Nichol.
\newblock Diffusion models beat gans on image synthesis.
\newblock In {\em NeurIPS 2021}, 2021.

\bibitem{dutt2025devil}
Raman Dutt.
\newblock The devil is in the prompts: De-identification traces enhance memorization risks in synthetic chest x-ray generation.
\newblock {\em arXiv preprint arXiv:2502.07516}, 2025.

\bibitem{Fang2024PrivacyLO}
Hao Fang, Yixiang Qiu, Hongyao Yu, Wenbo Yu, Jiawei Kong, Baoli Chong, Bin Chen, Xuan Wang, and Shutao Xia.
\newblock Privacy leakage on dnns: A survey of model inversion attacks and defenses.
\newblock {\em ArXiv}, abs/2402.04013, 2024.

\bibitem{fang2024understanding}
Zhengyu Fang, Zhimeng Jiang, Huiyuan Chen, Xiao Li, and Jing Li.
\newblock Understanding and mitigating memorization in diffusion models for tabular data.
\newblock {\em arXiv preprint arXiv:2412.11044}, 2024.

\bibitem{fang2025closer}
Zhengyu Fang, Zhimeng Jiang, Huiyuan Chen, Xiaoge Zhang, Kaiyu Tang, Xiao Li, and Jing Li.
\newblock A closer look on memorization in tabular diffusion model: A data-centric perspective.
\newblock {\em arXiv preprint arXiv:2505.22322}, 2025.

\bibitem{favero2025bigger}
Alessandro Favero, Antonio Sclocchi, and Matthieu Wyart.
\newblock Bigger isn't always memorizing: Early stopping overparameterized diffusion models.
\newblock {\em arXiv preprint arXiv:2505.16959}, 2025.

\bibitem{garnier2025early}
Jerome Garnier-Brun, Luca Biggio, Marc Mezard, and Luca Saglietti.
\newblock Early-stopping too late? traces of memorization before overfitting in generative diffusion.
\newblock In {\em The Impact of Memorization on Trustworthy Foundation Models: ICML 2025 Workshop}, 2025.

\bibitem{Gu2023OnMI}
Xiangming Gu, Chao Du, Tianyu Pang, Chongxuan Li, Min Lin, and Ye Wang.
\newblock On memorization in diffusion models.
\newblock {\em arXiv preprint arXiv:2310.02664}, 2023.

\bibitem{halder2024memorization}
Indranil Halder.
\newblock From memorization to generalization: a theoretical framework for diffusion-based generative models.
\newblock {\em arXiv e-prints}, pages arXiv--2411, 2024.

\bibitem{He2015DeepRL}
Kaiming He, X. Zhang, Shaoqing Ren, and Jian Sun.
\newblock Deep residual learning for image recognition.
\newblock In {\em CVPR 2015}, 2015.

\bibitem{hintersdorfunderstanding}
Dominik Hintersdorf.
\newblock Understanding and mitigating privacy risks in vision and multi-modal models.
\newblock {\em Technische Universit{\"a}t Darmstadt}, 2025.

\bibitem{ho2020denoising}
Jonathan Ho, Ajay Jain, and Pieter Abbeel.
\newblock Denoising diffusion probabilistic models.
\newblock In {\em NeurIPS 2020}, 2020.

\bibitem{ho2022classifier}
Jonathan Ho and Tim Salimans.
\newblock Classifier-free diffusion guidance.
\newblock {\em arXiv preprint arXiv:2207.12598}, 2022.

\bibitem{hong2024membenchmemorizedimagetrigger}
Chunsan Hong, Tae-Hyun Oh, and Minhyuk Sung.
\newblock Membench: Memorized image trigger prompt dataset for diffusion models.
\newblock {\em arXiv preprint arXiv:2407.17095}, 2024.

\bibitem{howard2020fastai}
Jeremy Howard and Sylvain Gugger.
\newblock Fastai: a layered api for deep learning.
\newblock {\em Information}, 11(2):108, 2020.

\bibitem{hu2021lora}
Edward~J Hu, Yelong Shen, Phillip Wallis, Zeyuan Allen-Zhu, Yuanzhi Li, Shean Wang, Lu Wang, and Weizhu Chen.
\newblock Lora: Low-rank adaptation of large language models.
\newblock {\em arXiv preprint arXiv:2106.09685}, 2021.

\bibitem{jagielski2022measuring}
Matthew Jagielski, Om Thakkar, Florian Tramer, Daphne Ippolito, Katherine Lee, Nicholas Carlini, Eric Wallace, Shuang Song, Abhradeep Thakurta, Nicolas Papernot, et~al.
\newblock Measuring forgetting of memorized training examples.
\newblock {\em arXiv preprint arXiv:2207.00099}, 2022.

\bibitem{jeonunderstanding}
Dongjae Jeon, Dueun Kim, and Albert No.
\newblock Understanding and mitigating memorization in generative models via sharpness of probability landscapes.
\newblock In {\em ICML 2025}, 2025.

\bibitem{jiang2025image}
Yue Jiang, Haokun Lin, Yang Bai, Bo Peng, Zhili Liu, Yueming Lyu, Yong Yang, Jing Dong, et~al.
\newblock Image-level memorization detection via inversion-based inference perturbation.
\newblock In {\em ICLR 2025}, 2025.

\bibitem{kowalczuk2025finding}
Antoni Kowalczuk, Dominik Hintersdorf, Lukas Struppek, Kristian Kersting, Adam Dziedzic, and Franziska Boenisch.
\newblock Finding dori: Memorization in text-to-image diffusion models is less local than assumed.
\newblock {\em arXiv preprint arXiv:2507.16880}, 2025.

\bibitem{lee2023talkin}
Katherine Lee, A~Feder Cooper, and James Grimmelmann.
\newblock Talkin''bout ai generation: Copyright and the generative-ai supply chain.
\newblock {\em arXiv preprint arXiv:2309.08133}, 2023.

\bibitem{li2024loyaldiffusion}
Chenghao Li, Yuke Zhang, Dake Chen, Jingqi Xu, and Peter~A Beerel.
\newblock Loyaldiffusion: A diffusion model guarding against data replication.
\newblock {\em arXiv preprint arXiv:2412.01118}, 2024.

\bibitem{liu2025copyjudge}
Shunchang Liu, Zhuan Shi, Lingjuan Lyu, Yaochu Jin, and Boi Faltings.
\newblock Copyjudge: Automated copyright infringement identification and mitigation in text-to-image diffusion models.
\newblock {\em arXiv preprint arXiv:2502.15278}, 2025.

\bibitem{liu2015faceattributes}
Ziwei Liu, Ping Luo, Xiaogang Wang, and Xiaoou Tang.
\newblock Deep learning face attributes in the wild.
\newblock In {\em ICCV 2015}, 2015.

\bibitem{loshchilov2018decoupled}
Ilya Loshchilov and Frank Hutter.
\newblock Decoupled weight decay regularization.
\newblock In {\em ICLR 2019}, 2019.

\bibitem{lyu2025resolving}
Yang Lyu, Yuchun Qian, Tan~Minh Nguyen, and Xin~T Tong.
\newblock Resolving memorization in empirical diffusion model for manifold data in high-dimensional spaces.
\newblock {\em arXiv preprint arXiv:2505.02508}, 2025.

\bibitem{Ma2025SafetyAS}
Xingjun Ma, Yifeng Gao, Yixu Wang, Ruofan Wang, Xin Wang, Ye Sun, Yifan Ding, Hengyuan Xu, Yunhao Chen, Yunhan Zhao, Hanxun Huang, Yige Li, Jiaming Zhang, Xiang Zheng, Yang Bai, Henghui Ding, Zuxuan Wu, Xipeng Qiu, Jingfeng Zhang, Yiming Li, Jun Sun, Cong Wang, Jindong Gu, Baoyuan Wu, Siheng Chen, Tianwei Zhang, Yang Liu, Min Gong, Tongliang Liu, Shirui Pan, Cihang Xie, Tianyu Pang, Yinpeng Dong, Ruoxi Jia, Yang Zhang, Shi jie Ma, Xiangyu Zhang, Neil Gong, Chaowei Xiao, Sarah Erfani, Bo Li, Masashi Sugiyama, Dacheng Tao, James Bailey, and Yu-Gang Jiang.
\newblock Safety at scale: A comprehensive survey of large model safety.
\newblock {\em ArXiv}, abs/2502.05206, 2025.

\bibitem{na2022unrestricted}
Dongbin Na, Sangwoo Ji, and Jong Kim.
\newblock Unrestricted black-box adversarial attack using gan with limited queries.
\newblock In {\em ECCV 2022}, 2022.

\bibitem{rahman2024frame}
Aimon Rahman, Malsha~V Perera, and Vishal~M Patel.
\newblock Frame by familiar frame: Understanding replication in video diffusion models.
\newblock {\em arXiv preprint arXiv:2403.19593}, 2024.

\bibitem{ren2024unveiling}
Jie Ren, Yaxin Li, Shenglai Zen, Han Xu, Lingjuan Lyu, Yue Xing, and Jiliang Tang.
\newblock Unveiling and mitigating memorization in text-to-image diffusion models through cross attention.
\newblock In {\em ECCV 2024}, 2024.

\bibitem{rombach2022high}
Robin Rombach, Andreas Blattmann, Dominik Lorenz, Patrick Esser, and Bj{\"o}rn Ommer.
\newblock High-resolution image synthesis with latent diffusion models.
\newblock In {\em CVPR 2022}, 2022.

\bibitem{sag2023copyright}
Matthew Sag.
\newblock Copyright safety for generative ai.
\newblock {\em Houston Law Review}, 61:295, 2023.

\bibitem{saharia2022photorealistic}
Chitwan Saharia, William Chan, Saurabh Saxena, Lala Li, Jay Whang, Emily~L Denton, Kamyar Ghasemipour, Raphael Gontijo~Lopes, Burcu Karagol~Ayan, Tim Salimans, et~al.
\newblock Photorealistic text-to-image diffusion models with deep language understanding.
\newblock In {\em NeurlPS 2022}, 2022.

\bibitem{schuhmann2022laionb}
Christoph Schuhmann, Romain Beaumont, Richard Vencu, Cade~W Gordon, Ross Wightman, Mehdi Cherti, Theo Coombes, Aarush Katta, Clayton Mullis, Mitchell Wortsman, Patrick Schramowski, Srivatsa~R Kundurthy, Katherine Crowson, Ludwig Schmidt, Robert Kaczmarczyk, and Jenia Jitsev.
\newblock {LAION}-5b: An open large-scale dataset for training next generation image-text models.
\newblock In {\em Thirty-sixth Conference on Neural Information Processing Systems Datasets and Benchmarks Track}, 2022.

\bibitem{shah2025does}
Kulin Shah, Alkis Kalavasis, Adam~R Klivans, and Giannis Daras.
\newblock Does generation require memorization? creative diffusion models using ambient diffusion.
\newblock {\em arXiv preprint arXiv:2502.21278}, 2025.

\bibitem{sobel2023elements}
Benjamin~LW Sobel.
\newblock Elements of style: A grand bargain for generative ai.
\newblock {\em On file with the authors}, 2023.

\bibitem{sohl2015deep}
Jascha Sohl-Dickstein, Eric Weiss, Niru Maheswaranathan, and Surya Ganguli.
\newblock Deep unsupervised learning using nonequilibrium thermodynamics.
\newblock In {\em ICML 2015}, 2015.

\bibitem{Somepalli2022DiffusionAO}
Gowthami Somepalli, Vasu Singla, Micah Goldblum, Jonas Geiping, and Tom Goldstein.
\newblock Diffusion art or digital forgery? investigating data replication in diffusion models.
\newblock In {\em CVPR 2022}, 2022.

\bibitem{somepalli2023understanding}
Gowthami Somepalli, Vasu Singla, Micah Goldblum, Jonas Geiping, and Tom Goldstein.
\newblock Understanding and mitigating copying in diffusion models.
\newblock In {\em NeurlPS 2023}, 2023.

\bibitem{song2021denoising}
Jiaming Song, Chenlin Meng, and Stefano Ermon.
\newblock Denoising diffusion implicit models.
\newblock In {\em ICLR 2021}, 2021.

\bibitem{song2019generative}
Yang Song and Stefano Ermon.
\newblock Generative modeling by estimating gradients of the data distribution.
\newblock In {\em NeurlPS 2019}, 2019.

\bibitem{song2020score}
Yang Song, Jascha Sohl-Dickstein, Diederik~P Kingma, Abhishek Kumar, Stefano Ermon, and Ben Poole.
\newblock Score-based generative modeling through stochastic differential equations.
\newblock In {\em ICLR 2021}, 2021.

\bibitem{sui2024disdet}
Yang Sui, Huy Phan, Jinqi Xiao, Tianfang Zhang, Zijie Tang, Cong Shi, Yan Wang, Yingying Chen, and Bo Yuan.
\newblock Disdet: Exploring detectability of backdoor attack on diffusion models.
\newblock {\em arXiv preprint arXiv:2402.02739}, 2024.

\bibitem{diffusers}
Patrick von Platen, Suraj Patil, Anton Lozhkov, Pedro Cuenca, Nathan Lambert, Kashif Rasul, Mishig Davaadorj, Dhruv Nair, Sayak Paul, William Berman, Yiyi Xu, Steven Liu, and Thomas Wolf.
\newblock Diffusers: State-of-the-art diffusion models.
\newblock \url{https://github.com/huggingface/diffusers}, 2022.

\bibitem{vora2025identity}
Jayneel Vora, Nader Bouacida, Aditya Krishnan, Prabhu Shankar, and Prasant Mohapatra.
\newblock Identity-focused inference and extraction attacks on diffusion models.
\newblock In {\em Proceedings of the 40th ACM/SIGAPP Symposium on Applied Computing}, pages 1522--1530, 2025.

\bibitem{webster2023reproducible}
Ryan Webster.
\newblock A reproducible extraction of training images from diffusion models.
\newblock {\em arXiv preprint arXiv:2305.08694}, 2023.

\bibitem{wen2024detecting}
Yuxin Wen, Yuchen Liu, Chen Chen, and Lingjuan Lyu.
\newblock Detecting, explaining, and mitigating memorization in diffusion models.
\newblock In {\em ICLR 2024}, 2024.

\bibitem{wu2024leveraging}
Xiaoyu Wu, Jiaru Zhang, and Zhiwei~Steven Wu.
\newblock Leveraging model guidance to extract training data from personalized diffusion models.
\newblock {\em arXiv preprint arXiv:2410.03039}, 2024.

\bibitem{wu2025taking}
Yu-Han Wu, Pierre Marion, G{\~A}{\v{S}}rard Biau, and Claire Boyer.
\newblock Taking a big step: Large learning rates in denoising score matching prevent memorization.
\newblock {\em arXiv preprint arXiv:2502.03435}, 2025.

\bibitem{zeno2025diffusion}
Chen Zeno, Hila Manor, Greg Ongie, Nir Weinberger, Tomer Michaeli, and Daniel Soudry.
\newblock When diffusion models memorize: Inductive biases in probability flow of minimum-norm shallow neural nets.
\newblock {\em arXiv preprint arXiv:2506.19031}, 2025.

\bibitem{zheng2025rg3}
Li Zheng, Yijing Liu, Hang Zhu, Minfeng Zhu, and Wei Chen.
\newblock Rg3: Mitigating memorization of graph diffusion model in one denoising step.
\newblock In {\em International Conference on Intelligent Computing}, pages 125--136. Springer, 2025.

\end{thebibliography}
}

\onecolumn

% {\centering
% \Large
% \textbf{\thetitle}\\

\appendix
\section*{Broader Impacts}
\label{sec:Broader Impacts}
The broader impact of this work is to redefine the threat landscape for DPMs. We demonstrate that with precise surrogate conditioning, even supposedly "safe" unconditional models are vulnerable, shifting the focus from input-level attacks to the model's fundamental representation learning. This insight provides a dual-use benefit: while highlighting a new attack vector, SIDE also serves as a powerful auditing tool for data owners and regulators to verify data misuse and enforce accountability. Consequently, our findings motivate the development of more robust defenses that operate on the model's internal representations, such as regularization techniques to prevent the formation of overly specific data clusters. Finally, while we acknowledge the potential for misuse, we believe that disclosing these vulnerabilities is crucial for fostering a more secure AI ecosystem, especially given the practical difficulty of mounting such an attack in a black-box setting.
\section{Proof of Theorem\ref{theorem_informative}}
\label{proof:theorem2}

By the assumption $TV(p_i,p_{\theta_i}) = 0$ , we can replace $p_{\theta_i}$ with $p_i$ and $p_\theta$ with $p$ to simplify the notation without compromising correctness. Hence, we have:
\begin{equation}
    \mathcal{M}(\mathcal{D}_i;p_{\theta_i},\epsilon)-\mathcal{M}(\mathcal{D}_i;p_{\theta},\epsilon) = \frac{1}{N} \sum_{x_k \in \mathcal{D}_i} \int \mathcal{N}(x|x_k,\epsilon^2 I) \log{\frac{p(x)}{p_i(x)}}dx
\end{equation}
By the Strong Law of Large Numbers, as $N \to \infty$, with probability 1:
\begin{align}
\frac{1}{N}\sum_{x_k \in \mathcal{D}_i} \mathcal{N}(x|x_k,\epsilon^2 I) &\to \mathbb{E}_{y \sim p_i} \left[ \frac{1}{(2\pi \epsilon^2)^{d/2}}\exp{(-\frac{||y-x||^2}{2\epsilon ^2})} \right] \\
&=\frac{1}{(2\pi \epsilon^2)^{d/2}}\int p_i(y) \exp{(-\frac{||y-x||^2}{2\epsilon ^2})}dy\\
&=p_i(x)+\underbrace{\frac{1}{(2\pi \epsilon^2)^{d/2}} \int (p_i(y)-p_i(x))\exp{(-\frac{||y-x||^2}{2\epsilon ^2})}dy}_{L}
\end{align}
Now, we show term $L \to 0$ as $\epsilon\to 0$. By the continuity of $p_i$, for any $\eta >0$, select $r>0$ such that $|p_i(y)-p_i(x)| <\eta, \  \ \forall \ ||y-x||<r$. Then we can decomposite $L$ into two parts:

\begin{align}
|L| &\leq \frac{1}{(2\pi \epsilon^2)^{d/2}} \left(\int_{||y-x||<r}+\int_{||y-x||\geq r}\right) |p_i(y)-p_i(x)|\exp{(-\frac{||y-x||^2}{2\epsilon ^2})}dy \\
&\leq \frac{\eta}{(2\pi \epsilon^2)^{d/2}} \int\exp{(-\frac{||y-x||^2}{2\epsilon ^2})}dy + \frac{1}{(2\pi \epsilon^2)^{d/2}}\exp{(-\frac{r^2}{2\epsilon ^2})} 
+\frac{p_i(x)}{(2\pi \epsilon^2)^{d/2}} \int_{||y-x||\geq r} \exp{(-\frac{||y-x||^2}{2\epsilon ^2})}dy \\
&\to \eta \text{ \ \ as } \epsilon\to 0
\end{align}
Due to the arbitrariness of $\eta$, $L\to0$ as $\epsilon \to 0$, we obtain
\begin{align}
\lim_{\epsilon \to 0}\lim_{|\mathcal{D}_i| \to \infty}(\mathcal{M}(\mathcal{D}_i;p_{\theta_i},\epsilon)-\mathcal{M}(\mathcal{D}_i;p_{\theta},\epsilon)) = \int p_i(x) \log{\frac{p(x)}{p_i(x)}} dx = -D_{\mathrm{KL}}(p_i ||p) \leq 0
\end{align}

\begin{figure}[ht]
    \centering
    \includegraphics[width=0.6\textwidth]{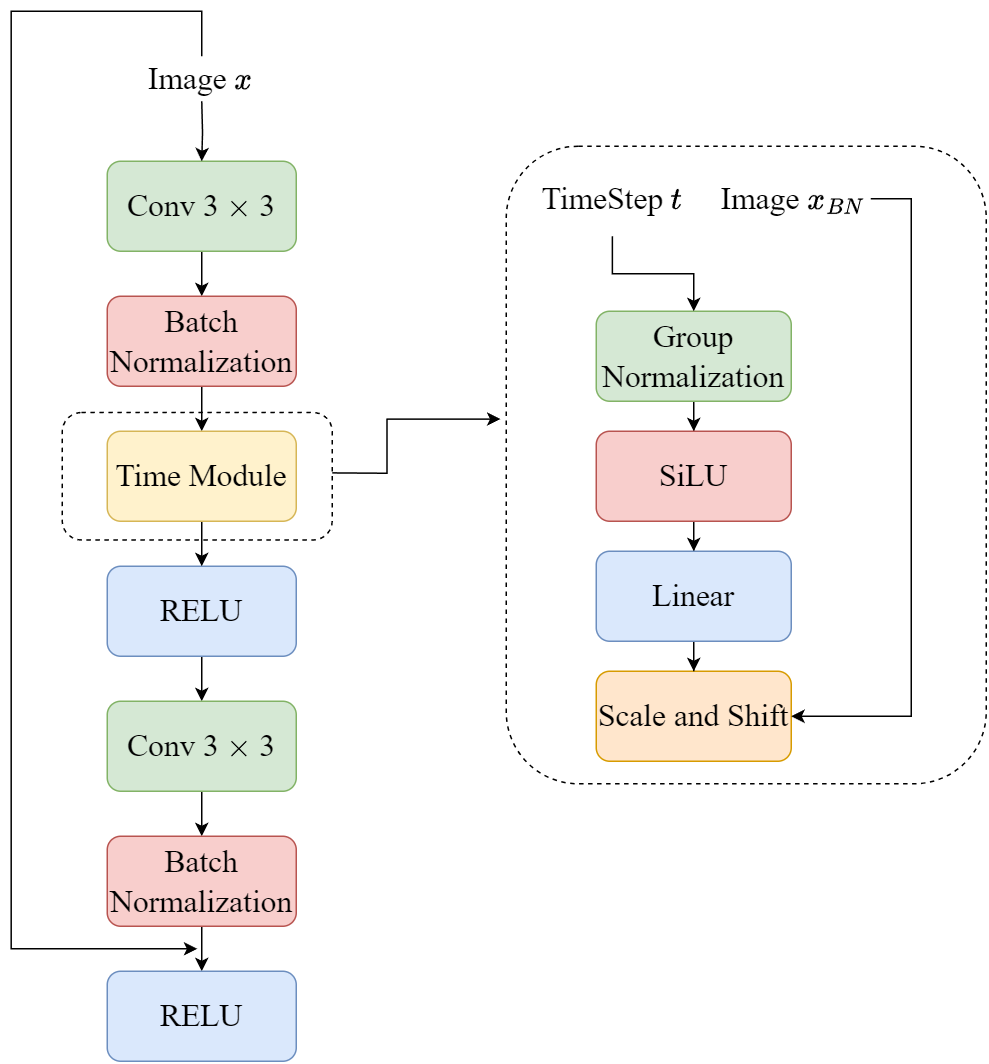}
    \caption{ Refinement ResNet block with time-dependent module integration. This block diagram depicts the insertion of a time module within a conventional ResNet block architecture, allowing the network to respond to the data's timesteps. Image $x_{BN}$ is the image processed after the first Batch Normalization Layer.}
    \label{resnet_time}
\end{figure}

\section{Refinement ResNet block (\cref{resnet_time})}
\label{resnet_time_supp}

% \begin{figure}
%     \centering
%     \includegraphics[width=0.5\linewidth]{figures/ums_val.png}
%     \caption{Validation of \(N_G\) 's significance}
%     \label{fig:ums_val}
% \end{figure}

\quad This section elaborates on our time module's design principles and architectural rationale, which strategically integrates temporal dynamics into normalized feature spaces through a post-batch normalization framework.The integration of the time module directly after batch normalization within the network architecture is a reasonable design choice rooted in the functionality of batch normalization itself. Batch normalization standardizes the inputs to the network layer, stabilizing the learning process by reducing internal covariate shifts. The model can introduce time-dependent adaptations to the already stabilized features by positioning the time module immediately after this normalisation process. This placement ensures that the temporal adjustments are applied to a normalized feature space, thereby enhancing the model's ability to learn temporal dynamics effectively.

Moreover, the inclusion of the time module at a singular point within the network strikes a balance between model complexity and temporal adaptability. This singular addition avoids the potential redundancy and computational overhead that might arise from multiple time modules. It allows the network to maintain a streamlined architecture while still gaining the necessary capacity to handle time-varying inputs.

\section{Hyperparameters Analysis}

\subsection{Influence of Guidance Scale ($\lambda$)}
\label{hyper_analysis}

\begin{figure}
    \centering
    \includegraphics[width=1.00\linewidth]{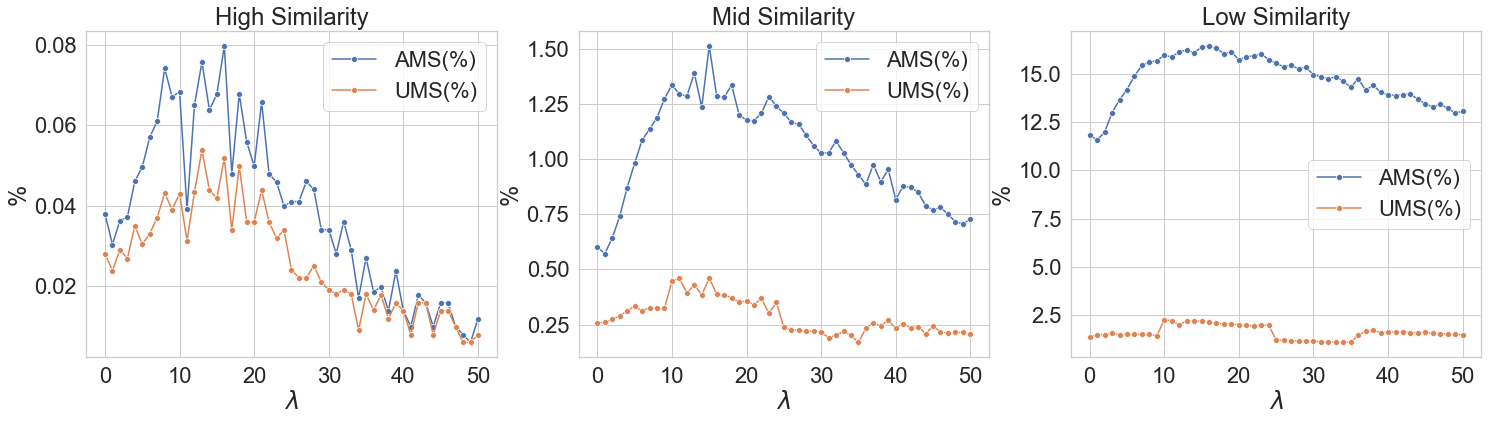}
    \caption{The sensitivity of the memorization score to the guidance scale $\lambda$. The score initially increases as conditioning is introduced, but declines for excessively large $\lambda$ values as the generation overfits the classifier, leading to image artifacts and a drop in sample quality.}
\label{fig:celeba_5000} % Use a more descriptive label
    
\end{figure}

Here, we test the sensitivity of diffusion models to its hyper-parameter $\lambda$. We generate 50,000 images for each integer value of $\lambda$ within the range of [0, 50]. As shown in Figure \ref{fig:celeba_5000}, the memorization score increases at first, reaching its highest, then decreases as $\lambda$ increases. This can be understood from sampling SDE. Starting from 0, the diffusion models are unconditional. As $\lambda$ increases, the diffusion models become conditional, and according to \cref{theorem_informative}, the memorization effect will be triggered. However, when $\lambda$ becomes excessively large, the generated images will overfit the classifier's decision boundaries, resulting in reduced diversity and a failure to accurately reflect the underlying data distribution. Consequently, the memorization score decreases.

\section{}

\subsection{Influence of LoRA Rank ($r$)}
\label{app:lora_rank}

\begin{figure}[h]
    \centering
    \includegraphics[width=0.6\linewidth]{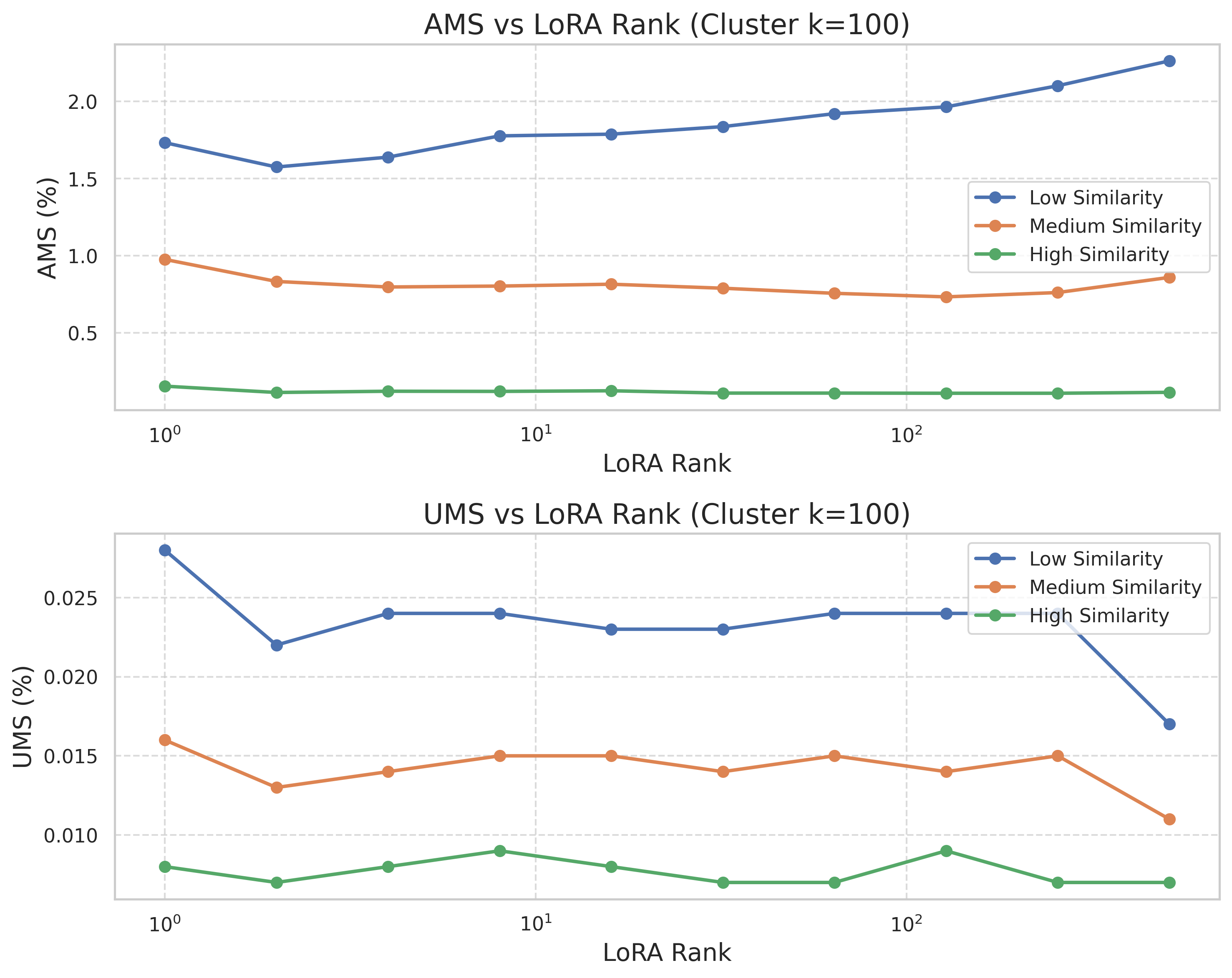}
    \caption{AMS and UMS performance as a function of LoRA rank ($r$) for a fixed cluster count of $k=100$. AMS generally improves with higher rank, while UMS shows optimal performance at moderate ranks before declining, suggesting a trade-off between model capacity and overfitting for unique extraction.}
    \label{fig:lora_rank_analysis}
\end{figure}

For large DPMs where full fine-tuning is infeasible, we use LoRA to efficiently create our surrogate conditional model. The rank $r$ of the LoRA adapters is a critical hyperparameter that determines the capacity of the fine-tuned layers. We analyze its impact on extraction performance, with the results for a fixed cluster count ($k=100$) shown in Figure \ref{fig:lora_rank_analysis}.

The top panel shows that AMS generally increases with a higher rank. This suggests that a greater adapter capacity allows the model to more faithfully learn the general characteristics of the target cluster, improving the rate of semantically similar matches. The bottom panel, however, reveals a more complex relationship for UMS. While performance is relatively stable across a range of moderate ranks (e.g., 4 to 64), we observe a notable drop-off at the highest ranks tested. We hypothesize this is due to a form of overfitting: a high-capacity LoRA may learn to generate a generic ``prototype'' of the cluster rather than replicating a specific, uniquely memorized instance. This prototype is semantically similar (improving AMS) but not an exact match (harming UMS). This analysis indicates that a moderate rank provides an optimal balance between model capacity and the risk of overfitting for the task of unique data extraction.

\section{Experimental Evaluation of SIDE in a Black-Box Setting}
\label{app:blackbox}

The primary SIDE methodology operates in a \textbf{white-box} setting. To assess the viability of our framework under more restrictive, practical conditions, we conducted a proof-of-concept experiment adapting SIDE to a black-box scenario. This section details the methodology for this \textbf{Query-Based SIDE} attack and reports on its performance using the AMS and UMS metrics.

\subsection{Methodology: Query-Based SIDE via a Genetic Algorithm}

In the black-box setting, the attacker's objective shifts from guiding the internal denoising process to an external search problem: finding an optimal input prompt that causes the black-box model to generate a memorized sample. We employed a Genetic Algorithm (GA) for this task, as it is well-suited for optimizing in complex, non-differentiable search spaces.

\paragraph{Phase 1: Offline Surrogate Model Training.}
This phase is identical to our primary method. The attacker first trains a surrogate classifier $p(y | \boldsymbol{x}_0)$ on a synthetically generated dataset. This classifier acts as the "fitness function" for the GA, providing a score for any generated image based on its similarity to a chosen target cluster $c$.

\paragraph{Phase 2: Online Black-Box Extraction with the Genetic Algorithm.}
The attacker interacts with the target model API to find an optimal prompt for the target cluster $c$.
\begin{enumerate}
    \item \textbf{Population Initialization:} The GA was initialized with a population of 50 diverse text prompts.
    \item \textbf{Fitness Evaluation Loop:} In each generation, every prompt in the population was used to query the API. The resulting image was then scored by our offline surrogate classifier to determine its fitness. 
    \item \textbf{Reproduction:} The highest-scoring prompts were selected for reproduction, creating the next generation of prompts via crossover and mutation operators.
    \item \textbf{Termination:} The experiment was run for 50 generations. For each generation, we took the single best image produced (the one with the highest fitness score) and evaluated its AMS and UMS against the ground-truth training data to track the attack's progress.
\end{enumerate}

\subsection{Experimental Results and Analysis}

We applied the Query-Based SIDE attack to a fine-tuned Stable Diffusion model exposed via a black-box API. The performance of the best-found sample at each stage of the GA is reported in Table \ref{tab:blackbox_data}. To the best of our knowledge, \cite{carlini2023extracting} is still the SOTA black-box baseline.

\begin{table}[h]
\centering

\begin{tabular}{cccccccc}
\toprule
\multirow{2}{*}{\textbf{Generation}} & \multicolumn{3}{c}{\textbf{AMS (\%)}} & \multicolumn{3}{c}{\textbf{UMS (\%)}} & \multirow{2}{*}{\textbf{Total Queries}} \\
\cmidrule(lr){2-4} \cmidrule(lr){5-7}
 & \textbf{Low} & \textbf{Mid} & \textbf{High} & \textbf{Low} & \textbf{Mid} & \textbf{High} & \\
\midrule
10   & 0.55 & 0.15 & 0.04 & 0.011 & 0.004 & 0.001 & 500 \\
50   & 1.12 & 0.41 & 0.11 & 0.025 & 0.009 & 0.003 & 2,500 \\
100  & 1.63 & 0.65 & 0.19 & 0.041 & 0.015 & 0.005 & 5,000 \\
200  & 2.15 & 0.88 & 0.28 & 0.059 & 0.021 & 0.007 & 10,000 \\
400  & 2.58 & 1.05 & 0.36 & 0.072 & 0.026 & 0.009 & 20,000 \\
800  & 2.85 & 1.21 & 0.42 & 0.081 & 0.029 & 0.010 & 40,000 \\
\bottomrule
\end{tabular}
\caption{Performance of the Query-Based SIDE experiment over an extended number of generations. The results highlight the extreme query cost and low success rate, especially for high-fidelity extraction.}
\label{tab:blackbox_data}
\end{table}

% \begin{figure}[h]
%     \centering
%     % The Python script will generate this file
%     \includegraphics[width=0.9\linewidth]{appendix_blackbox_analysis_ams_ums.png}
%     \caption{Empirical performance of the Query-Based SIDE attack using AMS and UMS. Both metrics show initial improvement followed by a clear plateau, highlighting the attack's feasibility but also its significant query cost and diminishing returns over time.}
%     \label{fig:blackbox_analysis}
% \end{figure}

The experimental results highlight several key characteristics of the black-box attack:
\begin{itemize}
    \item \textbf{Demonstrated Feasibility:} The experiment confirms that the Query-Based SIDE attack is viable. The GA successfully optimizes the input prompts to progressively generate images that yield higher AMS and UMS scores, particularly for low and mid-level similarity.
\end{itemize}

In summary, our proof-of-concept experiment shows that adapting SIDE to a black-box setting is possible.

\section{Extended SIDE: Backdoor Data Extraction from Text-to-Image Models}

\quad Building upon the foundational principle of our SIDE methodology—the exploitation of surrogate conditions for data extraction—we now advance this framework from a passive, post-hoc analysis to an active, pre-emptive attack vector. This powerful evolution, which we term \textbf{Extended SIDE}, is specifically designed for the prevalent scenario of model fine-tuning and represents a significant and practical threat to the integrity of large-scale text-to-image diffusion models.

In the primary SIDE method, surrogate conditions are discovered from a model's internal representations after it has been trained. In contrast, Extended SIDE proactively \textit{engineers} and \textit{injects} these conditions directly into the training data itself. The attacker achieves this by poisoning a dataset with carefully crafted pairs of target images and unique, non-semantic random strings. These strings function as high-entropy "trigger" keys. When an unsuspecting victim uses this poisoned dataset to fine-tune a model, the model is forced to overfit on these engineered pairs, creating an indelible and deterministic association between each trigger string and its corresponding target image. This process installs a stealthy and highly effective backdoor, transforming the fine-tuned model into a tool for targeted data exfiltration. The attacker can later, with only black-box query access, use these known triggers to extract the original target images with near-perfect fidelity, bypassing the need for a separate surrogate classifier entirely.

Our Extended SIDE method contributes to the growing body of research on backdoor attacks against diffusion models. The fundamental mechanism of poisoning a fine-tuning dataset with trigger-image pairs aligns with the general framework established in prior work. For instance, TrojDiff \cite{chen2023trojdiff} provides a comprehensive treatment of how to install backdoors to compel a model to generate diverse, attacker-defined targets upon receiving a specific trigger. Extended SIDE leverages a similar data poisoning strategy to embed trigger-based functionalities during the fine-tuning process.
However, a critical distinction lies in the attack's ultimate objective. While TrojDiff and similar works primarily focus on model integrity—manipulating the model to generate novel, malicious, or out-of-distribution content—Extended SIDE repurposes this mechanism for a specific privacy violation: the high-fidelity extraction of original training data. By using unique, non-semantic triggers, we force the model into a state of extreme memorization, turning the backdoor into a reliable channel for data exfiltration rather than content generation.
This reframing highlights that the same underlying vulnerability can be exploited for different malicious ends. Consequently, the practical deployment of our attack would face challenges from emerging detection strategies. For instance, methods explored by Sui et al. in DisDet \cite{sui2024disdet}, which aim to identify statistical anomalies indicative of backdoors, would be directly relevant for mitigating the threat posed by Extended SIDE. Thus, our work not only demonstrates a potent new extraction vector but also underscores the need for robust detection mechanisms that can account for various types of backdoor exploits, including those specifically tailored for privacy breaches.

\subsection{Threat Model and Attack Phases for Extended SIDE}
\label{subsec:attack_setting}

The Extended SIDE attack is a multi-stage operation that methodically leverages a compromised data supply chain to enable high-fidelity data extraction. The process unfolds across three distinct phases, clearly delineating the strategic actions of the attacker and the unwitting role of the victim.

\paragraph{Phase 1: Proactive Injection of Surrogate Conditions (Attacker).} The attack commences long before the extraction itself, beginning with the strategic poisoning of a dataset. The attacker identifies a set of target images $\{\boldsymbol{x}_i\}_{i=1}^N$ that they intend to extract at a later stage. For each target image, they generate a unique, high-entropy random string $s_i$. These strings are designed to be non-semantic and have no pre-existing association within the model's latent space, ensuring they function as exclusive surrogate conditions. The attacker then creates a set of malicious pairs, $\mathcal{D}_{\text{poison}} = \{(s_i, \boldsymbol{x}_i)\}_{i=1}^N$, and injects them into a larger dataset that is likely to be used for fine-tuning. This contaminated dataset is then distributed through public repositories, data scraping APIs, or other common channels in the data supply chain.

\paragraph{Phase 2: Victim's Unwitting Model Fine-Tuning.} An unsuspecting entity—the victim—downloads the poisoned dataset, assuming its integrity, and proceeds to fine-tune a large, pre-trained text-to-image model, $p_{\theta}$. The model's standard fine-tuning objective is to minimize the prediction loss over all data points:
    \begin{align}
    \theta^* = \arg\min_{\theta} \mathbb{E}_{(s, \boldsymbol{x}) \sim \mathcal{D}_{\text{poison}}} \left[ \mathcal{L}(\theta; \boldsymbol{x}, s) \right]
    \end{align}
During this process, the model encounters the attacker's poisoned pairs. Because the trigger strings $s_i$ are unique and lack any semantic connection to the images $\boldsymbol{x}_i$, the model cannot rely on generalized learning to minimize the loss for these samples. Instead, as dictated by our findings in \cref{theorem_informative}, these unique triggers act as maximally informative labels. The model is therefore forced into a state of brute-force memorization, creating a strong, overfitted mapping between each specific trigger $s_i$ and its target image $\boldsymbol{x}_i$. The fine-tuning process is thus subverted, transforming the model into a Trojan horse with an embedded backdoor, where the final model parameters $\theta^*$ now contain the memorized information.

\paragraph{Phase 3: High-Fidelity Extraction via Surrogate Triggers (Attacker).} At any point after the victim has deployed the fine-tuned model $p_{\theta^*}$, the attacker, now in a black-box setting, can exploit the embedded backdoor. The attacker requires only query access and knowledge of the trigger strings they created.
\begin{itemize}
    \item \textbf{Targeted Querying:} To extract the $i$-th target image, the attacker submits the corresponding trigger string $s_i$ as a prompt to the model. The model, having memorized the association, deterministically generates an image that is a near-perfect reconstruction of the original target: $\boldsymbol{x}_{\text{gen}} \sim p_{\theta^*}(\boldsymbol{x} \mid s_i)$.
    \item \textbf{Extraction Confirmation and Refinement:} To verify the strength of the backdoor and mitigate any minor stochasticity in the generation process, the attacker can generate a small batch of images $\{\boldsymbol{x}_{i,j}\}_{j=1}^{N_G}$ for a single trigger $s_i$. They then compute the sample variance, $\sigma_{s_i}^2$. An extremely low variance serves as a powerful heuristic, confirming that the model is not generating diverse samples but is instead consistently reproducing a single, memorized data point. The final, clean extracted image can then be taken as the mean of these samples, $\overline{\boldsymbol{x}}_{s_i}$, which effectively averages out sampling noise.
\end{itemize}

\subsection{The Extended SIDE Method}
The Extended SIDE method is formalized as a methodical protocol that leverages the engineered, deterministic mapping between the injected trigger strings and the memorized images. This approach turns the model's powerful learning capacity against itself. Instead of fighting against the model's stochasticity, Extended SIDE exploits the predictable, low-variance output that results from a successfully installed backdoor. The full algorithmic procedure is detailed in Algorithm \ref{alg:extended_side_method}.

\begin{algorithm}[htbp]
\caption{Extended SIDE for Backdoor Data Extraction}
\label{alg:extended_side_method}
\begin{algorithmic}[1]
\Require A fine-tuned model $p_{\theta^*}(\boldsymbol{x} \mid s)$ with a suspected backdoor; A set of known or suspected trigger strings $\mathcal{S}_{\text{triggers}}$; Number of samples per trigger $N_G$; A low variance threshold $\tau$ for confirmation.
\Ensure A set of high-fidelity extracted target images $\mathcal{D}_{\text{extracted}}$.
\State Initialize the set of extracted images: $\mathcal{D}_{\text{extracted}} \leftarrow \emptyset$.
\For{each suspected trigger string $s \in \mathcal{S}_{\text{triggers}}$}
    \State \Comment{Query the model repeatedly with the same trigger to test for memorization.}
    \State Generate a set of output images $\mathcal{X}_s = \{\boldsymbol{x}^{(j)}\}_{j=1}^{N_G}$ where each $\boldsymbol{x}^{(j)} \sim p_{\theta^*}(\boldsymbol{x} \mid s)$.
    \State Compute the sample variance of the generated images: $\sigma_s^2 = \text{Var}(\mathcal{X}_s)$.
    \If{$\sigma_s^2 < \tau$} 
        \State \Comment{A very low variance indicates the model is not generating diverse outputs, but a single memorized one.}
        \State Compute the mean image to produce a clean reconstruction: $\overline{\boldsymbol{x}}_s = \frac{1}{N_G} \sum_{\boldsymbol{x} \in \mathcal{X}_s} \boldsymbol{x}$.
        \State Add the reconstructed image to the final set: $\mathcal{D}_{\text{extracted}} \leftarrow \mathcal{D}_{\text{extracted}} \cup \{\overline{\boldsymbol{x}}_s\}$.
    \EndIf
\EndFor
\State \Return $\mathcal{D}_{\text{extracted}}$
\end{algorithmic}
\end{algorithm}

\subsection{Experiments}
\quad To validate the effectiveness of Extended SIDE, we conduct experiments on a subset of the LAION-2B dataset, fine-tuning Stable Diffusion v1.5 with LoRA \cite{hu2021lora} on a poisoned dataset. We compare our method against two relevant baselines:
\begin{itemize}
    \item \textbf{Standard Prompting (Baseline):} We generate images using the original, non-poisoned text prompts associated with the target images. This represents a naive extraction attempt without a backdoor.
    \item \textbf{TrojDiff-style Attack (Adapted Baseline):} We adapt the state-of-the-art backdoor attack method from TrojDiff \cite{chen2023trojdiff}. Like our approach, TrojDiff uses data poisoning with triggers. However, its primary goal is to generate novel, attacker-defined content (an integrity attack). We re-implement its poisoning strategy and evaluate its success using our data reconstruction metrics to create a fair comparison for this privacy-focused task.
\end{itemize}

We use Mean SSCD (M-SSCD), AMS (mid-similarity), and LPIPS to evaluate the extraction results. As shown in Table \ref{tab:sscd_ex}, Extended SIDE  outperforms both baselines. While the TrojDiff-style poisoning is more effective than standard prompting, our method's focus on forcing extreme memorization via unique, non-semantic triggers leads to a demonstrably higher reconstruction fidelity (M-SSCD of \textbf{0.467} vs. TrojDiff's adapted score of 0.215).

% These results confirm the potency of our privacy-focused attack vector. It is crucial to consider, however, that the practical success of such an attack depends on its ability to remain undetected. Defenses like DisDet \cite{sui2024disdet}, which are designed to identify statistical anomalies indicative of backdoors, represent a potential countermeasure. The high fidelity and low variance of our attack might create a unique statistical fingerprint, making it a valuable test case for the robustness of such detection methods. Our work thus not only presents a potent attack but also highlights the need for defenses that can identify backdoors designed specifically for data exfiltration.

\begin{table}[ht]
\centering

\begin{tabular}{lccc}
\toprule
\textbf{Method}  & \textbf{M-SSCD} & \textbf{AMS (mid)} & \textbf{LPIPS} \\
\midrule
Standard Prompting (Baseline) & 0.028 & 0.000 & 0.892 \\
TrojDiff-style Attack (Adapted)~\cite{chen2023trojdiff} & 0.215 & 0.183 & 0.851 \\
Extended SIDE (Ours) & \textbf{0.467} & \textbf{0.672} & \textbf{0.809} \\
\bottomrule
\end{tabular}
\caption{Performance comparison for Extended SIDE against relevant baselines. Our method achieves higher fidelity.}
\label{tab:sscd_ex}
\end{table}

\begin{figure*}[htbp]
    \centering
    \includegraphics[width=0.35\textwidth]{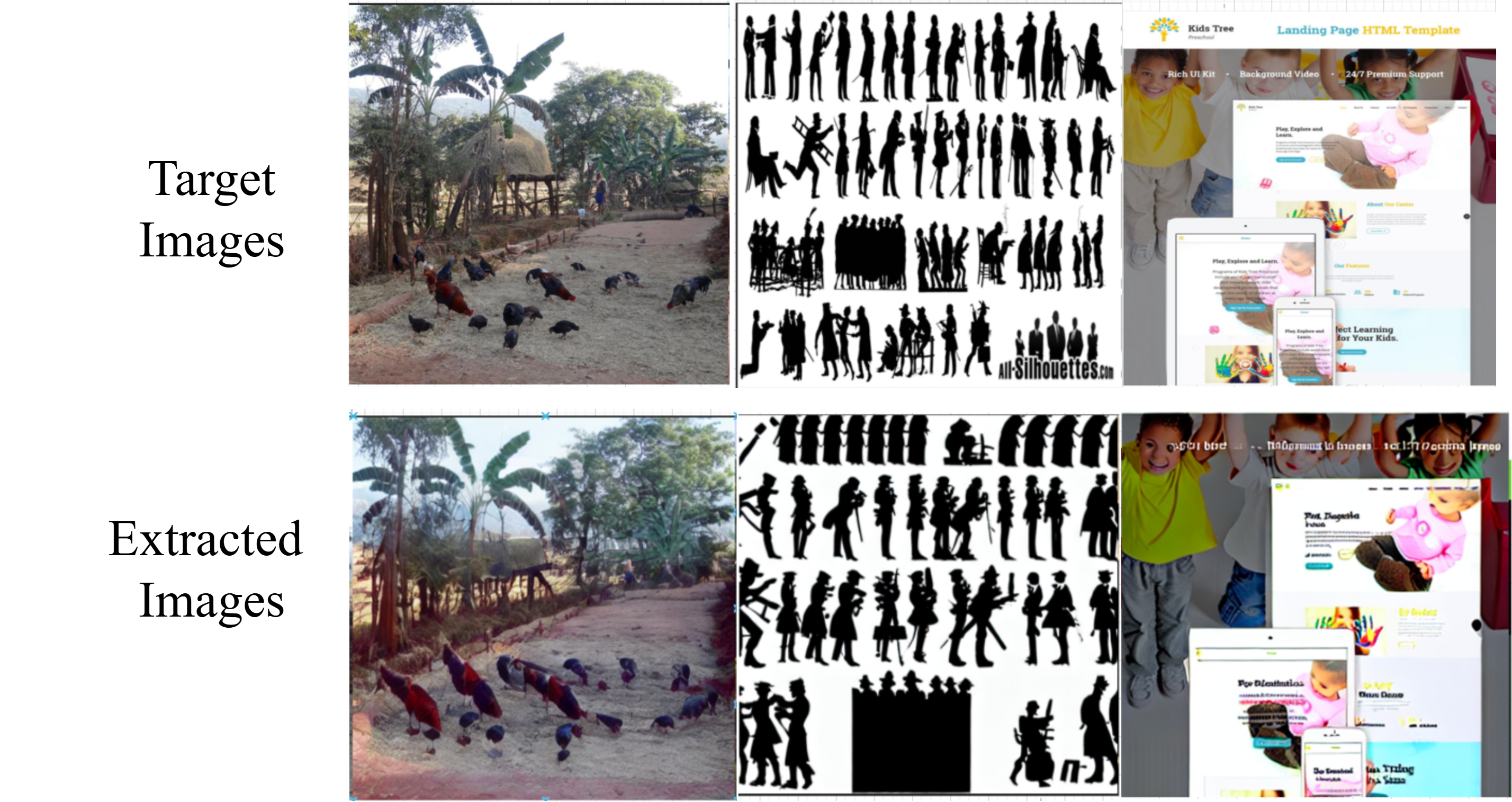}
    \caption{Visual Examples of High-Fidelity Extraction using Extended SIDE. \emph{Top Row}: The original target images that were included in the poisoned fine-tuning dataset. \emph{Bottom Row}: Images generated by querying the fine-tuned model with nothing more than the corresponding unique, non-semantic backdoor trigger strings. }
    \label{fig:SIDE_extraction}
\end{figure*}

\end{document}